\documentclass[11pt,a4paper]{article}

\usepackage{fontspec}
\usepackage{sortify-report}

\runningheader{Let the Agent Steer: Closed-Loop Ranking Optimization via Influence Exchange}

\begin{document}

\makereportheader
  {\includegraphics[height=28pt]{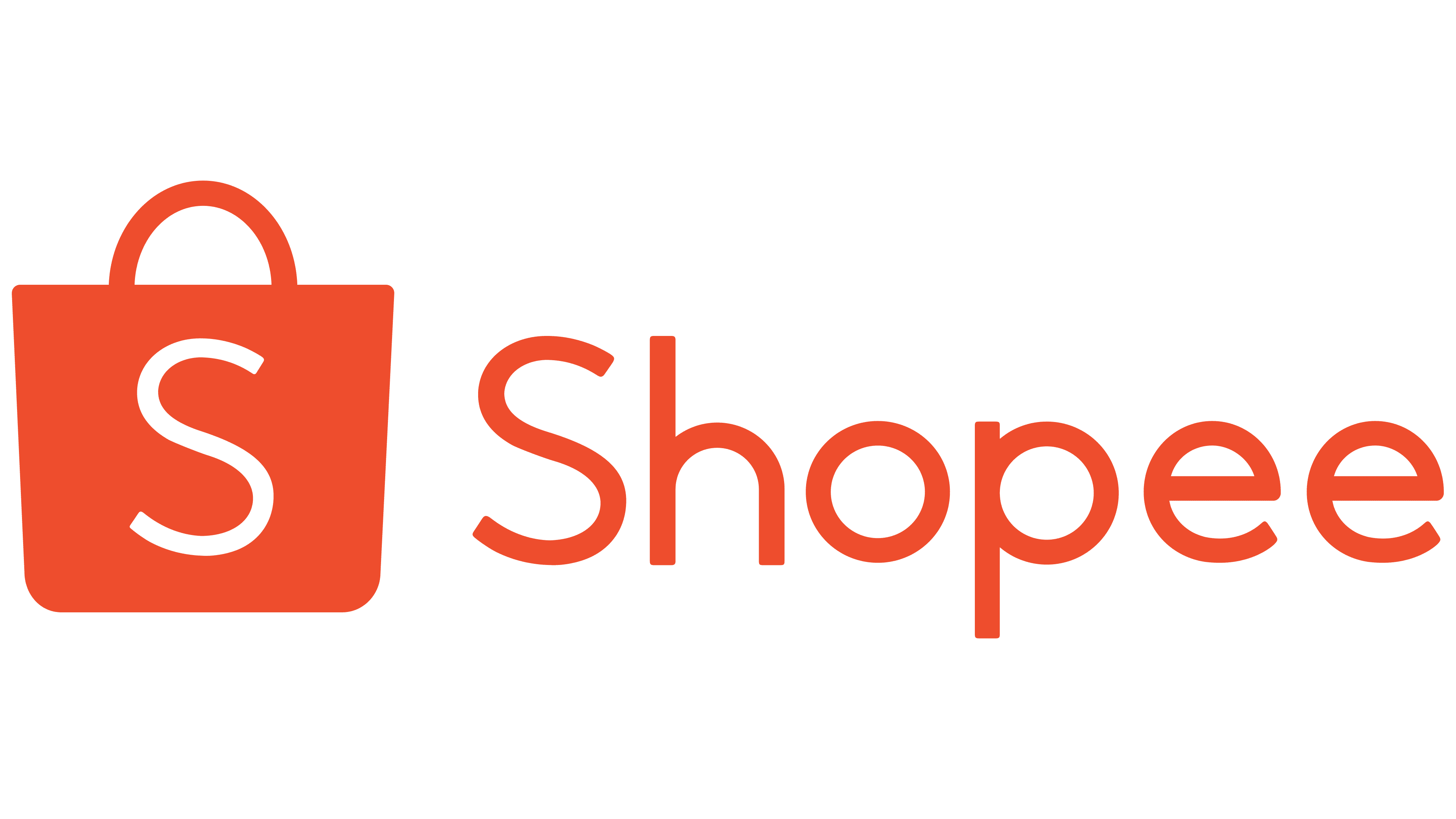}}
  {Sortify Technical Report}
  {March 2026}

\makereporttitle{Let the Agent Steer: Closed-Loop Ranking Optimization via Influence Exchange}

\makereportauthors{Sortify Team\footnote{Yin Cheng, Liao Zhou, Xiyu Liang, Dihao Luo, Tewei Lee, Kailun Zheng, Weiwei Zhang, Mingchen Cai, Jian Dong, Andy Zhang. All authors are from Shopee.}}


Recommendation ranking is fundamentally an influence allocation problem: a sorting formula distributes ranking influence among competing factors---organic relevance, advertising bids, price competitiveness---and the business outcome depends on finding the optimal ``exchange rates'' among them. However, offline proxy metrics systematically misjudge how influence reallocation translates to online impact: large offline uplifts often convert into only a small fraction of the expected online gains, with transfer rates far below expectation. More subtly, the bias pattern is asymmetric across metrics---some metrics exhibit optimistic bias (offline overestimates online), while others exhibit pessimistic bias (offline underestimates online)---so a single calibration factor cannot correct both. Traditional approaches worsen this predicament in three ways: manual calibration cannot keep pace with distribution shift; entangled diagnostic signals obscure whether the problem lies in mapping bias or constraint miscalibration; and each round starts from scratch, preventing historical experience from accumulating.

\figurePlaceholder{abs-overview}{Three-panel overview: (a) dual-channel architecture diagram, (b) online GMV/Orders uplift trend across rounds, (c) LLM correction convergence curve}

We present Sortify, a fully autonomous ranking optimization agent driven by a large language model (LLM). The agent reframes ranking optimization as continuous influence exchange, autonomously regulating the allocation of ranking influence among factors and closing the full loop from diagnosis and decision to parameter deployment without human intervention. It addresses the structural problems through three core mechanisms. First, grounded in L.J. Savage's axiomatization of Subjective Expected Utility (SEU)---which establishes that any rational decision requires exactly two independent inputs: a belief about the state of the world (probability) and a preference over outcomes (utility)---a dual-channel adaptation framework decouples offline-online transfer correction (the Belief channel) from constraint penalty adjustment (the Preference channel), architecturally severing the entanglement between epistemic error and axiological error to enable orthogonal diagnosis and independent correction. Second, an LLM meta-controller, serving as a second-order rational observer, operates on framework-level parameters---adjusting transfer function intercepts and penalty multipliers rather than low-level search parameters---and, based on evidence from 20-round episode histories, selectively corrects residuals left by routine LMS calibration. Third, a persistent Memory DB with 7 relational tables accumulates cross-round learning, providing the state basis for warm start and cross-round calibration continuity.

The agent's core evaluation metric, Influence Share, is a decomposable ranking influence measure in which all factor contributions sum to exactly 100\%. It addresses the fundamental limitation of traditional rank-correlation metrics such as Kendall's $\tau$, which cannot attribute influence to individual factors or support quantitative exchange between them. Sortify has been deployed on a large-scale recommendation platform spanning two Southeast Asian markets (hereafter Country~A and Country~B), running fully automated with no manual intervention required. The inner optimization loop operates autonomously in 4-hour cycles, executing 5,000 Optuna trials per round. In Country~A's warm-start experiment, the agent pushed GMV from $-$3.6\% to +9.2\% within 7 rounds, with peak order volume reaching +12.5\%, while LLM calibration effort converged from 5 correction items per round to 2. In Country~B, a cold-start deployment validated its cross-market generalization and end-to-end viability: after two search phases (23 rounds total), the agent identified a parameter structure that achieved +4.15\% GMV/UU and +3.58\% Ads Revenue in a 7-day A/B test, leading to a full production rollout. Overall, the evidence retained in the main text demonstrates that the system can both continuously improve on existing calibration state and produce deployable policies in new markets.

\vspace{12pt}

\newpage
\tableofcontents
\newpage


\section{Introduction}\label{sec:introduction}

Ranking in recommendation systems determines what users see first, directly impacting key business metrics such as gross merchandise value (GMV), order volume, advertising revenue, and user engagement. Industrial ranking stacks commonly combine multiple signals through tunable scoring functions and feature weights, especially in product search and marketplace recommendation settings~\cite{ltr_ecommerce_search,learning_to_rank_in_practice}. Optimizing these parameters is therefore a continuous, high-leverage engineering task. As platforms scale across markets, verticals, and recommendation surfaces, the need for automated, self-correcting parameter optimization becomes increasingly urgent~\cite{automl}.

\begin{figure}[H]
  \centering
  \includegraphics[width=0.9\textwidth]{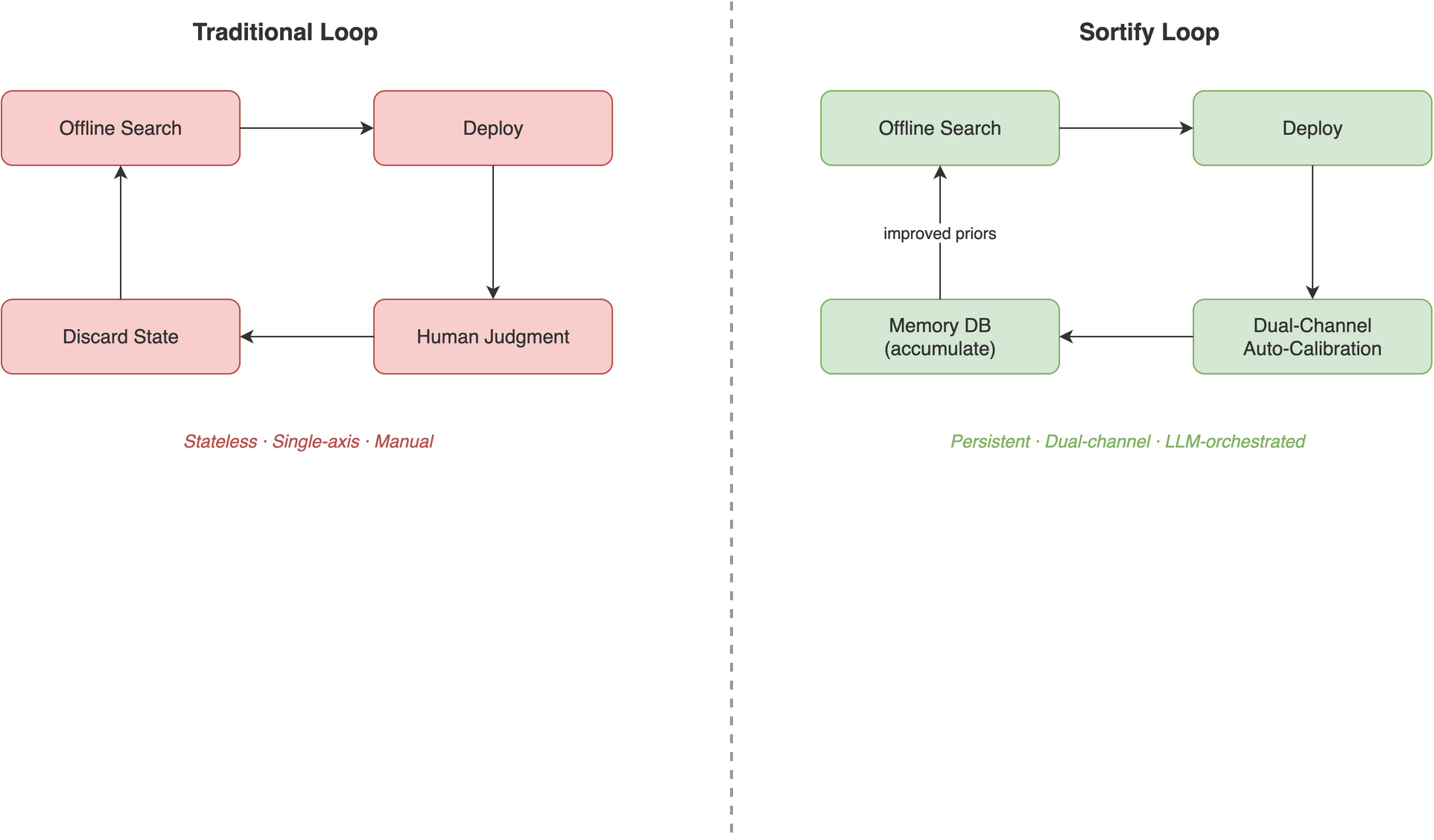}
  \caption{Old vs.\ New Paradigm: Left---manual optimization loop (offline search, deploy, human judgment, discard state, repeat). Right---Sortify closed loop (offline search, deploy, auto-calibrate via dual channels, accumulate in Memory DB, repeat with improved priors). Highlights the structural differences: stateless vs.\ persistent, single-axis vs.\ dual-channel, manual vs.\ LLM-orchestrated.}
  \label{fig:intro-paradigm}
\end{figure}

The standard optimization workflow operates as follows: an offline search algorithm explores the parameter space using proxy metrics computed on logged data, identifies a candidate configuration, and pushes it to production for online A/B validation. A human operator then examines the A/B results, judges whether the outcome is acceptable, and either adopts the new parameters or reverts to the previous configuration~\cite{counterfactual_evaluation}. This manual loop, while conceptually simple, suffers from three structural limitations that incremental tuning improvements cannot resolve.

\textbf{Offline-Online Transfer Gap.}
The most critical structural problem is the systematic divergence between offline proxy metrics and online business outcomes. In the Country~A warm-start experiment retained in the main text, offline $I_\text{gmv}$ rose from +18.2\% to +41.6\%, yet the corresponding online GMV fluctuated from $-$3.6\% to +9.2\%, revealing a significant optimistic bias in the offline proxy toward online gains. More critically, the same round's parameter update affects GMV, Orders, and Ads Revenue non-synchronously, indicating that the transfer relationships across different business metrics do not share a single calibration constant. Standard logged-data evaluation methods estimate policy value from historical interactions~\cite{counterfactual_evaluation,offline_policy_evaluation}, but they do not by themselves provide continuously updated, per-metric transfer calibration for a non-stationary production environment. Our production observations show that the mapping drifts continuously with traffic patterns, time-of-day effects, and competitive dynamics. Without continuous recalibration, each carefully designed offline search is built on potentially stale mapping assumptions, and the precision gains from search are partially offset by transfer error.

\textbf{Entangled Diagnostic Signals.}
When an optimization round produces disappointing online results, the operator faces a fundamental attribution problem: did the offline-online mapping predict incorrectly (a \textit{Belief} error), or were the constraint penalties calibrated at the wrong sensitivity level (a \textit{Preference} error)? These two failure modes require opposite corrections---a Belief error demands adjusting the transfer function intercept, while a Preference error demands rescaling penalty weights~\cite{bayesian_decision_theory}. Without formal separation, the two signal sources are conflated: a round where GMV is positive but Ads Revenue is negative could equally mean the mapping is optimistic or the constraint is too loose. This diagnostic entanglement is not merely a parameter tuning issue; it exposes a limitation of monolithic optimization workflows that compress multiple objectives and constraints into a single correction channel~\cite{pareto_ecommerce_recommendation}.

\textbf{No Persistent Learning.}
Each optimization round in conventional systems starts from a blank slate. The transfer relationships learned in round~$N$ are not carried forward to round~$N+1$. The constraint sensitivities discovered after a violation are forgotten after the next parameter push. This ``Groundhog Day'' effect means that a system having completed 50 rounds of A/B tests possesses no more calibration intelligence than one running its first round~\cite{continual_learning}. In our production setting, each offline-online observation pair represents 3.5~hours of live traffic data---a non-trivial investment. Discarding this accumulated evidence is both economically wasteful and technically avoidable. Recent work on continual learning~\cite{continual_learning}, meta-learning for warm-started optimization~\cite{meta_learning_hpo}, and LLM-based agents with persistent state~\cite{llm_agents_optimization} suggests that cumulative learning across episodes is both feasible and valuable for sequential decision problems.

We present Sortify, an agent-steered closed-loop system that reframes sorting parameter optimization as continuous influence exchange. Rather than treating optimization as a stateless manual process, Sortify enables an LLM agent to steer the rebalancing of ranking influence among competing factors through a persistent, autonomous feedback loop. Sortify's core architectural innovation is the decomposition of the influence exchange problem into two orthogonal channels---Belief (how influence predictions translate to reality) and Preference (how hard influence constraints are enforced)---orchestrated by an LLM meta-controller that operates on framework-level parameters rather than bottom-level search parameters.

The design of Sortify rests on three principles. First, \textit{separation of concerns}: transfer mapping correction (Belief) and constraint sensitivity adjustment (Preference) operate on independent axes derived from a Bayesian decision-theoretic decomposition $P(\theta|D) \times L(a,\theta)$~\cite{bayesian_decision_theory}, ensuring that corrections along one axis do not interfere with the other. Second, \textit{framework-level LLM control}: the LLM adjusts the search framework---specifically, transfer function intercepts and penalty multipliers---rather than the 7-dimensional sorting parameter vector directly. This distinction matters because framework parameters encode reusable cross-round experience (e.g., ``offline GMV predictions have been 5--8\% optimistic over the past 3 rounds''), while bottom-level parameters are ephemeral facts tied to a single data snapshot~\cite{meta_learning_hpo}. Third, \textit{cumulative memory}: a 7-table persistent Memory DB stores every offline-online observation, calibration update, and LLM proposal, enabling the system to build calibration intelligence across rounds rather than resetting after each iteration~\cite{continual_learning,meta_learning_hpo}.

The contributions of this report are as follows:

\begin{itemize}
  \item \textbf{Dual-channel self-adaptation framework.} We propose a Belief/Preference decomposition that enables orthogonal calibration of offline-online transfer mapping and constraint sensitivity. In warm-start production A/B tests on a Country~A recommendation platform, this framework pushes GMV from \textbf{$-$3.6\%} to \textbf{+9.2\%} within 7 rounds, with peak order volume reaching \textbf{+12.5\%} (Section~\ref{sec:online-ab}).

  \item \textbf{LLM meta-controller with evidence-based reasoning.} We introduce an LLM agent that adjusts two framework-level knobs---transfer function intercept ($\delta_\text{intercept} \in [-0.1, +0.1]$) and penalty multiplier ($\in [0.5, 2.0]$)---grounded in 20-round episode histories and 30-update calibration traces. In the Country~A warm-start experiment, the LLM's calibration effort converges from 5 correction items per round to 2 within 7 rounds, indicating that framework-level corrections diminish as evidence accumulates (Section~\ref{sec:llm-calibration}).

  \item \textbf{Persistent memory architecture for cumulative learning.} We demonstrate that a 7-table relational Memory DB enables the system to warm-start on existing calibration state and extend cross-round learning into subsequent experiments. The Country~A case starts directly from an existing Memory DB state, while the Country~B case completes a cross-market deployment without architectural modification (Sections~\ref{sec:parameter-stability} and~\ref{sec:vietnam}).
\end{itemize}

The remainder of this report is organized as follows. Section~\ref{sec:architecture} presents the system architecture, detailing the Influence Share metric, dual-channel mechanism, LLM meta-controller, and Memory DB design. Section~\ref{sec:operational} describes the operational framework, including the 10-step one-shot pipeline and YOLO continuous loop with failure recovery. Section~\ref{sec:evaluation} provides evaluation evidence from 30 optimization rounds across Country~A and Country~B, comprising 7 Country~A warm-start rounds and 23 Country~B deployment rounds. Section~\ref{sec:conclusion} discusses limitations---including residual parameter oscillation, statistical uncertainty in low-traffic windows, and insufficient evidence for extreme structural breaks---and outlines future directions.

\newpage
\section{System Architecture}
\label{sec:architecture}

In this section, we present the complete architecture of Sortify, a three-layer closed-loop system for continuous sorting parameter optimization. The system takes online A/B experiment data as input, calibrates its internal model of offline-online transfer relationships, and outputs optimized sorting parameters for the next deployment cycle. We detail the five core components: the Influence Share metric and parameter search engine (Section~\ref{subsec:influence-share}), the Belief channel for transfer calibration (Section~\ref{subsec:belief-channel}), the Preference channel for constraint adaptation (Section~\ref{subsec:preference-channel}), the LLM meta-controller (Section~\ref{subsec:llm-meta-controller}), and the persistent Memory DB (Section~\ref{subsec:memory-db}).

\subsection{Overview}
\label{subsec:arch-overview}

Sortify operates through a three-layer architecture that separates concerns across human configuration, intelligent calibration, and parameter search.

\begin{itemize}
  \item \textbf{Layer 1 (Outer --- Human/Configuration):} Defines optimization objectives (e.g., maximize $I_\text{gmv}$), constraint boundaries (e.g., $I_\text{order}$ degradation $\leq$ 5\%), initial parameter ranges, and penalty weight seeds. This layer is set once per market and updated infrequently.

  \item \textbf{Layer 2 (Middle --- LLM + Algorithm):} Performs dual-channel calibration. The Belief channel corrects the offline-to-online transfer mapping via LMS regression and LLM-driven intercept adjustments. The Preference channel adapts constraint penalty weights via multiplicative updates. The LLM meta-controller orchestrates both channels using evidence from accumulated episode history.

  \item \textbf{Layer 3 (Inner --- Optuna TPE Search):} Executes 5,000 trials with 25 parallel workers in the 7-dimensional parameter space, operating under the calibrated target ranges and penalty weights produced by Layer~2.
\end{itemize}

The dual-channel design in Layer 2 is not an arbitrary engineering partition but a necessary consequence of rational decision axioms. L.J. Savage's~(1954) theory of Subjective Expected Utility (SEU) establishes that \textbf{any rational decision requires exactly two independent inputs: what you believe the world to be (probability/belief) and what you care about (utility/preference).} If a system's behavior satisfies consistency axioms, its preferences can be represented by subjective probabilities and a utility function; the utility representation is unique only up to a positive affine transformation. Sortify operationalizes this axiom into a system-level decomposition:

\begin{itemize}
  \item \textbf{Belief channel (truth-seeking):} Corresponds to the \textit{state belief} in Savage's theory---``How do offline metrics map to online reality?'' It concerns only objective physics and system regularities, carrying no value judgment.
  \item \textbf{Preference channel (value-seeking):} Corresponds to the \textit{utility function} in Savage's theory---``How much does it hurt when a constraint is violated?'' It defines only loss boundaries, without interfering with physical regularities.
\end{itemize}

If these two are not forcibly decoupled, the system falls into unavoidable cognitive biases: \textbf{Wishful Thinking}---distorting the objective assessment of offline-online transfer rates because of an intense desire for goal achievement; or \textbf{Sour Grapes}---reducing the importance of red-line constraints to make the overall loss function converge when transfer rates are poor. The dual-channel design severs this ``diagnostic entanglement'' at the architectural level, ensuring independent self-calibration along two orthogonal directions.

Furthermore, an intelligent agent can err in two orthogonal directions, each requiring a mutually exclusive correction path:

\begin{enumerate}
  \item \textbf{Epistemic Error:} ``I predicted it would happen, but it didn't''---the system's world model has failed. The correction \textbf{must not} modify penalty weights; it must correct the belief mapping.
  \item \textbf{Axiological Error:} ``It happened, but I didn't expect it to hurt this much''---the prediction was correct but the utility assessment was insufficient. The correction \textbf{must not} modify the transfer mapping; it must correct the utility function.
\end{enumerate}

The complete data flow forms a closed loop: online A/B data enters the Memory DB as a new episode $\to$ LMS updates transfer model slopes and intercepts $\to$ LLM proposes framework-level corrections $\to$ calibrated constraints feed into Optuna search $\to$ best parameters are published to Redis $\to$ new A/B data is generated $\to$ the cycle repeats. Each cycle completes in approximately 4 hours.

\figurePlaceholder{arch-overview}{Three-layer architecture diagram. Layer 1 (Human/Config) feeds objectives and constraints to Layer 2 (LLM + Algorithm), which contains the Belief channel, Preference channel, and LLM meta-controller. Layer 2 outputs calibrated target\_range and penalty\_weight to Layer 3 (Optuna TPE Search, 5000 trials $\times$ 25 workers). Layer 3 produces best parameters $\to$ Redis $\to$ online A/B $\to$ Memory DB $\to$ back to Layer 2.}

\subsection{Influence Share and Parameter Search Engine}
\label{subsec:influence-share}

\textbf{Physical Intuition: Rooms and Walls in High-Dimensional Space}

Before diving into the detailed mathematical derivation and the engineering motivation behind the system, let us establish an intuitive geometric and physical analogy to understand what ``ranking'' and ``influence'' truly represent.

Imagine a search request that recalls $n$ candidate items. From a traditional engineering perspective, ranking simply means calculating a total score for each of these $n$ items and ordering them from highest to lowest. However, if we shift our perspective to a high-dimensional geometric space, a completely different and more profound picture emerges.

We can view the scores of these $n$ items as a single ``point'' (a score vector) in an $n$-dimensional space. Within this vast space, there exist invisible ``walls''. When do we hit a wall? The critical state where any two items have exactly identical scores forms a ``hyperplane wall'' (a comparison boundary) in this space. Because there are $n$ items, the space is intersected by numerous such walls.

These intersecting walls partition the entire $n$-dimensional space into multiple closed ``rooms'' (geometrically known as chambers). Every single ``room'' represents a specific, deterministic, and unique final ranking order. As long as our score vector---the ``point''---stays safely inside a particular room, no items have tied scores, and the final ranking list remains absolutely fixed.

Now, consider the task of optimizing the ranking algorithm. In Sortify, we do this by adjusting the parameters $\boldsymbol{\theta}$ (such as changing the weight of GMV), which alters the final scores. In our geometric analogy, adjusting the parameters is essentially \textbf{pushing this ``point'' to move continuously through the space}.

As the parameters change, the point moves. If it merely wanders within the boundaries of a single room, the ranking order does not change at all, even though the underlying scores are shifting. It is only when the point \textbf{crosses a ``wall''}---stepping from one room into an adjacent one---that two items swap their ranks, and the overall ranking list actually changes.

This brings us to the core insight of our system: \textbf{the atomic event of ranking change is not the reshuffling of a global list, but the point crossing a specific wall.}

Following this logic, how should we define the \textbf{``influence''} of individual business factors (such as orders, GMV, or ads) on the final ranking?

Since a rank change is equivalent to ``the point crashing through a wall,'' we can simply perform a physical ``force analysis.'' The distance to the wall is determined by the score difference between two items, and this total score difference is the sum of differences contributed by individual factors. Therefore, each factor acts as a ``pusher'' exerting force in a specific direction. The ``influence'' of each factor is determined by \textbf{how much ``force'' it exerts in the normal direction of that wall} to push the point across.

The \textit{Influence Share} metric measures exactly this: the percentage of pushing force each factor contributes relative to the total push. By adopting this physical intuition, we translate abstract parameter tuning and permutation-group dynamics into a crystal-clear force and kinetic attribution analysis in high-dimensional space. This perspective liberates us from opaque rank correlation metrics (like Kendall Tau) and allows us to delve into the microscopic physical process of every rank swap to precisely quantify the contribution of each business factor.

\subsubsection{Motivation: Beyond Kendall Tau}
\label{subsubsec:beyond-kendall-tau}

Traditional parameter search in sorting systems relies on rank correlation metrics such as Kendall Tau to measure how well a candidate parameter set preserves a target ranking order. While Kendall Tau captures ordinal agreement, it has two critical limitations for multi-factor sorting optimization. First, it is \textit{not decomposable}: Kendall Tau produces a single scalar that cannot be attributed to individual factors, making it impossible to answer questions like ``how much of the sorting decision is driven by GMV versus order count?'' Second, it does \textit{not support factor trade-off analysis}: because Kendall Tau has no sum-to-one constraint, there is no natural way to express ``shift 5\% of sorting influence from orders to GMV while keeping ad exposure stable.''

\subsubsection{Influence Share Definition: From Symmetric Groups to the Influence Permutation Algorithm}
\label{subsubsec:influence-share-definition}

We do not treat Influence Share as an isolated heuristic definition. Instead, we derive it as a continuous attribution scheme built on the permutation structure of ranking itself. Consider a request $q$ with $n_q$ items, indexed by $X_q=\{1,\dots,n_q\}$. Under parameter vector $\boldsymbol{\theta}$, the ranking outcome for that request is a permutation in the symmetric group:

\begin{equation}
\pi_q(\boldsymbol{\theta}) \in S_{n_q}.
\label{eq:arch-15}
\end{equation}

Here $\pi_q(\boldsymbol{\theta})(k)$ denotes the item placed at rank $k$. Since the symmetric group $S_{n_q}$ is generated by adjacent transpositions $s_k=(k,k+1)$, any difference between two rankings can be decomposed into a sequence of local swaps. For ranking control, the atomic event is therefore not ``replacing one permutation by another'' but flipping the relative order of a specific item pair.

To connect this discrete group structure to continuous parameters, let candidate parameters $\boldsymbol{\theta}$ induce a total score vector $\mathbf{S}_q(\boldsymbol{\theta}) \in \mathbb{R}^{n_q}$ for request $q$. For any $i \neq j$, define the comparison hyperplane and the corresponding chamber:

\begin{equation}
H_{qij} := \{x \in \mathbb{R}^{n_q} : x_i = x_j\}, \qquad
C_{\pi} := \{x : x_{\pi(1)} > x_{\pi(2)} > \cdots > x_{\pi(n_q)}\}.
\label{eq:arch-16}
\end{equation}

When $\mathbf{S}_q(\boldsymbol{\theta}) \in C_{\pi}$, the ranking is exactly permutation $\pi$; when it crosses some $H_{qij}$, the relative order of items $i$ and $j$ flips. This is the classical reflection-geometric representation of the symmetric group. The local coordinate governing ranking change is therefore the pairwise margin:

To avoid notational ambiguity, we assume scores are in generic position; if ties occur, a fixed deterministic tie-breaker resolves them, so the final ranking can still be treated as an element of $S_{n_q}$.

\begin{equation}
\Delta_q(i,j;\boldsymbol{\theta}) := S_q(i;\boldsymbol{\theta}) - S_q(j;\boldsymbol{\theta}).
\label{eq:arch-17}
\end{equation}

In Sortify, the total score is additively decomposed over factors:

\begin{equation}
S_q(i;\boldsymbol{\theta}) = \sum_{f \in \mathcal{F}} \text{score}_{q,f}(i;\boldsymbol{\theta}),
\qquad
\Delta_q(i,j;\boldsymbol{\theta}) = \sum_{f \in \mathcal{F}} \Delta_{q,f}(i,j;\boldsymbol{\theta}),
\label{eq:arch-18}
\end{equation}

where
\begin{equation}
\Delta_{q,f}(i,j;\boldsymbol{\theta}) := \text{score}_{q,f}(i;\boldsymbol{\theta}) - \text{score}_{q,f}(j;\boldsymbol{\theta}).
\end{equation}

This gives the key bridge from group theory to algorithm design: parameter updates do not act on a discrete permutation directly. They first reshape the margin of each pair along the normal direction of the comparison wall; the sign of that margin determines whether a swap occurs; and the accumulation of such swaps yields a new permutation. Under this lens, parameter search becomes an influence permutation algorithm: it continuously reallocates each factor's share of responsibility for potential swaps across important item pairs.

To turn that responsibility into a stable metric, we impose five attribution requirements at the pair level: locality, sign invariance, non-negativity, partition of unity, and invariance under common positive rescaling. Under these requirements, the simplest choice is an $L_1$ normalization of absolute factor margins. Define the total pairwise influence budget as:

\begin{equation}
Z_{qij}(\boldsymbol{\theta}) := \sum_{f \in \mathcal{F}} \left|\Delta_{q,f}(i,j;\boldsymbol{\theta})\right|.
\label{eq:arch-19}
\end{equation}

If $Z_{qij}(\boldsymbol{\theta}) = 0$, then no factor provides any discriminative signal for that pair, so the pair carries no ranking information and can be excluded from aggregation. For all informative pairs, the pairwise influence share of factor $f$ is defined as:

\begin{equation}
s_{q,f}(i,j;\boldsymbol{\theta}) = \frac{\left|\Delta_{q,f}(i,j;\boldsymbol{\theta})\right|}{Z_{qij}(\boldsymbol{\theta})}.
\label{eq:arch-01}
\end{equation}

It follows immediately that $\sum_{f \in \mathcal{F}} s_{q,f}(i,j;\boldsymbol{\theta}) = 1$. The role of the absolute value is to separate direction from magnitude: the sign of $\Delta_{q,f}$ indicates which item the factor pushes forward, while $\left|\Delta_{q,f}\right|$ measures how strongly it pushes. Holding other components fixed, $s_{q,f}(i,j;\boldsymbol{\theta})$ increases monotonically with $\left|\Delta_{q,f}(i,j;\boldsymbol{\theta})\right|$, giving an interpretable local influence signal without requiring the global permutation to vary monotonically with every parameter.

To move from pair-level shares to a request-level or dataset-level metric, we aggregate only informative and business-relevant pairs. Let $P_q$ denote the pair set for request $q$; it may contain all informative pairs or be restricted to a Top-$(K+L)$ region. For $(i,j)\in P_q$, let $r_{qi}$ denote the rank of item $i$ in request $q$, and define the position-sensitive weight:

\begin{equation}
w_{qij} = \exp\!\Big(-\frac{\min(r_{qi}, r_{qj})}{\tau}\Big).
\label{eq:arch-02}
\end{equation}

Here $\tau$ controls the decay rate. Pairs involving higher-ranked items receive larger weights, reflecting the business reality that top-of-page positions matter disproportionately.

The aggregate Influence Share for factor $f$ is then:

\begin{equation}
I_f(\boldsymbol{\theta}) = \frac{\sum_q \sum_{(i,j)\in P_q} w_{qij} \cdot s_{q,f}(i,j;\boldsymbol{\theta})}{\sum_q \sum_{(i,j)\in P_q} w_{qij}}.
\label{eq:arch-03}
\end{equation}

By construction, $\sum_{f \in \mathcal{F}} I_f(\boldsymbol{\theta}) = 1$. This sum-to-one property is what enables ``exchange'' reasoning: increasing $I_\text{gmv}$ by 5 percentage points necessarily decreases other factors by a combined 5 percentage points, making trade-offs explicit and quantifiable. At this point the derivation closes: a ranking in $S_{n_q}$ is generated by pairwise swaps, swaps are triggered by pairwise margins crossing comparison walls, and Influence Share measures how much of those potential swaps is attributable to each factor. Operationally, the influence permutation algorithm consists of five steps: compute factor-wise scores and sum them into total scores; sort to obtain $\pi_q(\boldsymbol{\theta})$; compute $\Delta_{q,f}$ and $s_{q,f}$ on $P_q$; aggregate with $w_{qij}$; and pass $I_f(\boldsymbol{\theta})$ to the downstream objective and constraints.

\subsubsection{Multi-Objective Search Formulation}
\label{subsubsec:multi-objective-search}

The parameter search optimizes a composite objective that maximizes the primary business metric ($I_\text{gmv}$ uplift) while penalizing constraint violations via quadratic penalties:

\begin{equation}
\mathcal{J}(\boldsymbol{\theta}) = 10 \cdot \underbrace{\frac{I_\text{gmv}(\boldsymbol{\theta}) - I_\text{gmv}^\text{base}}{|I_\text{gmv}^\text{base}|}}_{\text{relative uplift}} \;-\; \sum_{j=1}^{J} \lambda_j \cdot \big[\max(0,\; v_j(\boldsymbol{\theta}))\big]^2
\label{eq:arch-04}
\end{equation}

where $\boldsymbol{\theta} = (\text{ps\_ads\_wo}, \text{ps\_ads\_wg}, \text{ps\_org\_wo}, \text{ps\_org\_wg}, \text{ps\_porg\_w}, \text{ps\_price\_pow}, \text{ps\_w2})$ is the 7-dimensional parameter vector, $v_j(\boldsymbol{\theta})$ is the violation of the $j$-th constraint (positive when violated), and $\lambda_j$ is the penalty weight for constraint $j$.

The quadratic penalty form is a deliberate design choice:

\begin{equation}
\text{penalty}_j = \lambda_j \cdot v_j^2
\label{eq:arch-05}
\end{equation}

A 1\% violation incurs a penalty of $\lambda_j \times 0.01^2$, while a 10\% violation incurs $\lambda_j \times 0.10^2$ --- a 100x difference. This convexity tolerates minor boundary brushes while aggressively penalizing large violations, biasing the search toward solutions where all constraints are mildly satisfied rather than one being severely violated. Linear penalties, by contrast, would treat a 10\% violation as merely 10x worse than a 1\% violation, insufficient for industrial constraint enforcement.

Constraints include bounds on $I_\text{order}$ degradation, $I_\text{ecpm\_term}$ degradation, ads top-10 exposure rate, and $I_\text{gmv\_ads}$. The penalty weights $\lambda_j$ are initialized at values between 15,000 and 30,000 and are subsequently adapted by the Preference channel (Section~\ref{subsec:preference-channel}).

\figurePlaceholder{arch-influence-share}{Flow diagram showing: item pairs $\to$ pairwise score differences $\to$ per-factor share (sum=1) $\to$ rank-weighted aggregation $\to$ $I_\text{gmv}$, $I_\text{order}$, $I_\text{ecpm\_term}$. Below: 7 parameters map into the sorting formula, changing factor contributions.}

\subsubsection{Search Engine}
\label{subsubsec:search-engine}

The search is executed by Optuna's TPE (Tree-structured Parzen Estimator) sampler \cite{optuna}, which models the conditional distribution of parameters given good versus bad objective values. Each round runs \textbf{5,000 trials} with \textbf{25 parallel workers} using the Constant Liar strategy --- when a worker starts a trial before others finish, it assumes pending trials will return a pessimistic value, preventing redundant exploration. A typical search completes in \textbf{15--30 minutes}.

As demonstrated in Section~\ref{sec:online-ab}, this search engine consistently finds parameter configurations that achieve $I_\text{gmv}$ uplifts of +4.9\% to +41.6\% relative to baseline, subject to calibrated constraints.

\subsection{Belief Channel: Offline-Online Transfer Calibration}
\label{subsec:belief-channel}

\subsubsection{Motivation}
\label{subsubsec:belief-motivation}

The offline-online transfer gap documented in Section~\ref{sec:introduction} is not a fixed constant --- it varies across metrics, drifts over time, and shifts abruptly with traffic pattern changes. A static calibration (e.g., ``offline GMV predictions are always 3x too optimistic'') would become stale within days. The Belief channel addresses this by maintaining a \textit{continuously updated} linear transfer model for each metric pair, combining a slow algorithmic estimator (LMS) with a fast pattern-recognizing agent (LLM).

From a Bayesian epistemological perspective, the Belief channel embodies two fundamentally different philosophies for confronting an uncertain world---a dual tempo of belief updating: \textbf{normal science with smooth priors} (LMS) represents the classical Bayesian framework, assuming the world is continuous (Markovian), with each round of new evidence incrementally refining the prior; \textbf{paradigm shift and black swans} (LLM intercept jumps) represent the decisive recognition of structural breakpoints when the underlying distribution undergoes cliff-like changes---discarding the old prior and forcibly injecting a new one. The former handles routine distributional drift; the latter responds to sudden paradigmatic disruptions. Together, they constitute a complete response to an uncertain world.

\subsubsection{Linear Transfer Model}
\label{subsubsec:linear-transfer-model}

For each metric pair (e.g., $I_\text{gmv} \to$ GMV, $I_\text{order} \to$ Orders), we model the offline-to-online relationship as:

\begin{equation}
\hat{u}_\text{online} = \alpha \cdot u_\text{offline} + \beta
\label{eq:arch-06}
\end{equation}

where $u_\text{offline}$ and $u_\text{online}$ are the observed offline and online uplifts respectively, $\alpha$ (slope) captures the transfer rate, and $\beta$ (intercept) captures the systematic bias. Six such relationships are maintained: GMV\textasciitilde$I_\text{gmv}$, Orders\textasciitilde$I_\text{order}$, Ads Revenue\textasciitilde$I_\text{ecpm\_term}$, and three derived metrics.

\subsubsection{LMS Continuous Calibration}
\label{subsubsec:lms-calibration}

After each round, the slope and intercept are updated via Least Mean Squares (LMS) regression with learning rate $\eta = 0.2$:

\begin{align}
\alpha_{t+1} &= \alpha_t + \eta \cdot e_t \cdot u_{\text{offline},t} \label{eq:arch-07} \\
\beta_{t+1} &= \beta_t + \eta \cdot e_t \label{eq:arch-08}
\end{align}

where the prediction error $e_t = u_{\text{online},t} - \hat{u}_{\text{online},t}$. LMS provides stable, monotonic convergence for gradual distribution drift. However, under sudden structural shifts (e.g., a promotional event or traffic source change), LMS with $\eta = 0.2$ requires approximately \textbf{15 rounds} to reach 95\% convergence --- equivalent to over 2 days of production data, during which the system operates with a miscalibrated transfer model.

\subsubsection{LLM Selective Jumps}
\label{subsubsec:llm-selective-jumps}

To compress recovery time from structural breaks, the LLM meta-controller (Section~\ref{subsec:llm-meta-controller}) can directly adjust the intercept $\beta$ in discrete jumps of up to $\pm 0.1$:

\begin{equation}
\beta_\text{new} = \beta_\text{old} + \Delta\beta_\text{LLM}, \quad \Delta\beta_\text{LLM} \in \{0, \pm 0.05, \pm 0.1\}
\label{eq:arch-09}
\end{equation}

The LLM examines 20-round episode history and can detect patterns invisible to single-observation LMS --- for example, ``5 consecutive rounds where GMV transfer was pessimistic despite positive $I_\text{gmv}$ trends.'' When the LLM detects such a pattern, a single intercept jump can accomplish what LMS would need 10+ rounds to achieve. Critically, the LLM adjusts \textit{only the intercept}, not the slope, because intercept shifts represent systematic bias changes while slope changes indicate fundamental relationship restructuring, which should be left to the more conservative LMS estimator.

The output of the Belief channel is a calibrated \textbf{target\_range} for each constrained metric --- the offline uplift range that, after applying the calibrated transfer function, maps to an acceptable online outcome.

\figurePlaceholder{arch-dual-channel}{Two-axis diagram. Horizontal axis: Belief (target\_range = position of constraint boundary). Vertical axis: Preference (penalty\_weight = hardness of constraint boundary). Arrows show LMS moving along Belief axis continuously, LLM making discrete jumps along Belief axis, and violation pressure moving along Preference axis. The two axes are orthogonal --- corrections on one do not affect the other.}

As shown in Section~\ref{sec:transfer-analysis}, even under warm-start conditions the Belief channel must continue updating with new observations: in Exp-401838, R2 still exhibited a clearly optimistic offline bias, while by R7 the same kind of positive offline signal had become able to translate into significantly positive online GMV.

\subsection{Preference Channel: Constraint Penalty Adaptation}
\label{subsec:preference-channel}

\subsubsection{Motivation}
\label{subsubsec:preference-motivation}

Even with a perfectly calibrated transfer model, the search framework still needs to know \textit{how hard} to enforce each constraint. A constraint that was harmlessly brushed in round $N$ may become critically binding in round $N+1$ due to changing market conditions. The Preference channel continuously adjusts constraint penalty weights based on observed violation patterns.

Within Savage's SEU framework, the Preference channel addresses a fundamentally different error type from the epistemic error corrected by the Belief channel---\textbf{axiological error}: ``It happened, but I didn't expect it to hurt this much.'' The system's prediction was entirely correct, but its assessment of the outcome's ``pain'' (utility loss) was severely insufficient. For instance, the system accurately predicted a minor decline in a metric, the online result confirmed a minor decline, yet this small decline triggered a catastrophic alert in the business dashboard. In this case, the correction path \textbf{must not touch the transfer mapping}---the world model is not wrong---but must instead correct the utility function: the system is learning that ``this red line is far harder than expected; next time, even at the cost of the primary objective, it must never be crossed again.''

\subsubsection{Violation Pressure}
\label{subsubsec:violation-pressure}

For each constraint $j$, we compute a normalized violation pressure:

\begin{equation}
p_j = \frac{v_j(\boldsymbol{\theta})}{v_j^\text{threshold}}
\label{eq:arch-10}
\end{equation}

where $v_j(\boldsymbol{\theta})$ is the observed violation magnitude and $v_j^\text{threshold}$ is the acceptable violation threshold. A pressure of $p_j > 0$ indicates the constraint is being violated; $p_j \leq 0$ indicates it is satisfied.

\subsubsection{Asymmetric Multiplicative Update}
\label{subsubsec:asymmetric-update}

The penalty weight update follows a multiplicative rule with asymmetric step sizes:

\begin{align}
\delta_j &= \begin{cases} \delta_\text{up} = 0.25 & \text{if } p_j > 0 \\ -\delta_\text{down} = -0.08 & \text{if } p_j \leq 0 \end{cases} \label{eq:arch-11} \\[6pt]
\lambda_j^{(t+1)} &= \lambda_j^{(t)} \cdot \exp(\delta_j) \label{eq:arch-12}
\end{align}

This design encodes three deliberate properties:

\begin{enumerate}
  \item \textbf{Scale invariance.} Multiplicative updates apply equal \textit{percentage} changes regardless of the absolute magnitude of $\lambda_j$. When one constraint has $\lambda = 1{,}000$ and another has $\lambda = 80{,}000$, equal violation pressure produces proportionally equal adjustments. Additive updates would over-correct small weights and under-correct large ones.

  \item \textbf{Loss aversion.} The asymmetry $\delta_\text{up} = 0.25 > \delta_\text{down} = 0.08$ means that tightening a constraint (in response to violation) is \textbf{3x faster} than relaxing it (when satisfied). This reflects the industrial reality that the cost of a constraint violation (e.g., ad revenue collapse) outweighs the opportunity cost of an overly tight constraint (e.g., slightly suboptimal GMV).

  \item \textbf{Log-space interpretability.} Because the update is multiplicative in the original space and additive in log-space, $\delta_\text{up}$ and $\delta_\text{down}$ are directly interpretable as one-round tightening and relaxation steps. They are intentionally asymmetric, making the system more responsive to violations than to relaxation opportunities.
\end{enumerate}

The Preference channel's output is a set of calibrated \textbf{penalty\_weight} values that feed directly into the objective function (Eq.~\ref{eq:arch-04}). As demonstrated in Section~\ref{sec:parameter-stability}, the Preference channel's adaptive weights contribute to narrowing parameter oscillation amplitude across successive experiments.

\subsection{LLM Meta-Controller}
\label{subsec:llm-meta-controller}

\subsubsection{Motivation}
\label{subsubsec:llm-motivation}

The LMS estimator in the Belief channel handles routine distribution drift effectively but is fundamentally a single-observation, fixed-step-size algorithm. It cannot detect multi-round patterns (e.g., ``5 consecutive rounds of pessimistic GMV transfer''), distinguish between noise and structural breaks, or reason about the \textit{relationship} between Belief corrections and Preference corrections. The LLM meta-controller fills this gap by providing episodic pattern recognition and cross-channel reasoning.

From a cybernetics perspective, the LLM meta-controller achieves a leap from first-order control to \textbf{second-order control (reflection)}. First-order control---the Optuna search engine finding optimal solutions in the high-dimensional parameter space under given Belief and Preference conditions---is direct control over the ``environment.'' The LLM meta-controller, by contrast, does not directly manipulate the low-level sorting parameters $\boldsymbol{\theta}$. Its inputs are ``how the system has been perceiving data'' (LMS update histories and observation errors in the Memory DB)---it is \textbf{learning how the system learns}. This intervention on framework-level meta-parameters (intercepts and penalty multipliers) endows the system with the capacity for self-reflection. What the LLM accomplishes is not parameter optimization, but ``belief epiphany'' and ``value recalibration.''

\subsubsection{Two Orthogonal Knobs}
\label{subsubsec:two-knobs}

The LLM operates on exactly two output variables:

\begin{align}
\Delta\beta_\text{LLM} &\in [-0.1, +0.1] \quad \text{(Belief channel: intercept adjustment)} \label{eq:arch-13} \\
m_\text{LLM} &\in [0.5, 2.0] \quad \text{(Preference channel: penalty multiplier)} \label{eq:arch-14}
\end{align}

These bounds are safety constraints, not suggestions. A single-round intercept shift of $\pm 0.1$ is already a large correction; the cap prevents catastrophic miscalibration from a single LLM error. The penalty multiplier range of $[0.5, 2.0]$ allows halving or doubling a constraint's enforcement strength, sufficient for responding to structural breaks without enabling runaway penalty inflation.

A critical design decision is that the LLM adjusts \textit{framework-level} parameters (intercepts and penalty multipliers), not the 7-dimensional bottom-level sorting parameter vector ($\boldsymbol{\theta}$). The rationale is grounded in the distinction between \textbf{experience} and \textbf{facts}: ``offline GMV predictions have been 5--8\% optimistic over the past 3 rounds'' is a reusable structural insight that transfers across data snapshots, while ``ps\_ads\_wg = 2.15'' is a one-time fact about today's data distribution that may be 1.73 tomorrow. The LLM operates in the space of reusable experience; Optuna operates in the space of ephemeral facts.

\figurePlaceholder{arch-llm-controller}{Input assembly: 20-round episode history (offline/online metric pairs) + 30 prior update records (LMS and LLM correction history) + current penalty weights $\to$ Context JSON $\to$ LLM reasoning $\to$ Output: \{delta\_intercept, penalty\_multiplier, reason, evidence\_keys\}. Two arrows from output: delta\_intercept feeds into Belief channel (intercept update), penalty\_multiplier feeds into Preference channel (weight scaling). Safety bounds shown as guard rails on both outputs.}

\subsubsection{Context and Evidence Requirements}
\label{subsubsec:context-evidence}

The LLM receives a structured context JSON containing: (1) the current slope and intercept for all 6 prior relations, (2) the 20 most recent episodes with paired offline-online metrics, (3) the 30 most recent prior update history records (both LMS and LLM), and (4) the current penalty weight state.

Every LLM proposal must cite specific episode keys as evidence. Proposals without evidence are rejected. When the LLM's confidence is low --- for instance, if the recent episodes show contradictory signals --- it is instructed to return an empty proposal (no adjustment), which is preferable to a weakly grounded correction.

\subsubsection{Implementation: Model, Prompt, and Output Safety}
\label{subsubsec:llm-implementation}

\paragraph{Model and invocation.}
The meta-controller invokes an LLM through the Codex CLI's\footnote{OpenAI Codex CLI, \url{https://github.com/openai/codex}} \texttt{exec} subcommand in a sandboxed subprocess. Sortify implements its own memory system (Section~\ref{subsec:memory-db}) and tool-calling mechanisms, making it a fully capable autonomous agent in its own right---Codex serves solely as the underlying LLM inference engine, not as an agent framework. The invocation command is:

\begin{verbatim}
codex exec - --skip-git-repo-check \
  --sandbox read-only --color never \
  -c model_reasoning_effort="high"
\end{verbatim}

The prompt is passed through standard input, and the response is read from standard output. The reasoning effort is set to high, which activates extended reasoning before the model produces its structured JSON output. A 300-second timeout prevents pipeline stalls. No temperature parameter is set explicitly; the extended-reasoning mode subsumes sampling control by directing the model to reason thoroughly before committing to an answer.

\paragraph{Prompt structure.}
The LLM receives a single-shot prompt consisting of two concatenated parts:

\begin{enumerate}
  \item \textbf{System instructions} ($\sim$600 words, hardcoded): Define the output JSON schema (\texttt{proposal\_id}, an array of \texttt{proposals} each containing \texttt{relation\_key}, \texttt{delta\_intercept}, \texttt{penalty\_multiplier}, \texttt{reason}, and \texttt{evidence\_episode\_keys}), specify safety bounds ($|\Delta\beta| \leq 0.1$, $m \in [0.5, 2.0]$), and frame the task as Bayesian decision alignment---separating belief updates (intercept corrections on the transfer-function) from preference updates (loss-weight scaling for constraint violations).

  \item \textbf{Context JSON} (dynamically generated): The structured context described in Section~\ref{subsubsec:context-evidence}, exported by the \texttt{export-llm-context} command from the Memory DB.
\end{enumerate}

No few-shot examples are provided. The prompt is deliberately kept zero-shot to avoid anchoring the model on past correction patterns, which could suppress novel structural insights.

\paragraph{Output parsing and safety layers.}
The LLM's raw text output passes through a seven-layer safety pipeline before any system state is modified:

\begin{enumerate}
  \item \textbf{Prompt-level instruction:} The system instructions explicitly constrain output format and value ranges.
  \item \textbf{Robust JSON extraction:} A three-strategy parser attempts direct \texttt{json.loads}, then Markdown code-block extraction via regex, then a brace-depth state machine to isolate the first complete JSON object from potentially noisy output.
  \item \textbf{Schema validation and type coercion:} Each proposal field is validated: missing \texttt{delta\_intercept} defaults to 0.0, missing \texttt{penalty\_multiplier} defaults to 1.0, and all numeric fields are cast to \texttt{float}. Malformed entries are silently dropped.
  \item \textbf{Relation key whitelist:} Every \texttt{relation\_key} in the proposal is checked against the set of keys present in the context's prior relations. Unknown keys trigger an immediate rejection.
  \item \textbf{Hard clamp:} Regardless of the LLM's output, $\Delta\beta_\text{LLM}$ is clamped to $[-0.1, +0.1]$ and $m_\text{LLM}$ to $[0.5, 2.0]$ in code.
  \item \textbf{Global weight bounds:} Even after multiplicative scaling, penalty weights stay within configured floors and ceilings (e.g., $[1{,}000, 80{,}000]$), preventing cumulative drift across rounds.
  \item \textbf{Failure fallback:} If the LLM call times out, returns unparseable output, or produces an empty proposal array, the system falls back to a no-op proposal ($\Delta\beta = 0$, $m = 1.0$). The pipeline continues without interruption.
\end{enumerate}

This defense-in-depth design ensures that the LLM is a \textit{bounded advisor}: it can accelerate convergence through structural insight, but cannot destabilize the system even under adversarial or degenerate outputs.

\subsubsection{Coordination with LMS}
\label{subsubsec:coordination-lms}

LMS and LLM operate in serial within each round: LMS updates first (Step 5 of the pipeline), then the LLM reads the post-LMS state and proposes adjustments (Steps 7--8). This ordering ensures the LLM operates on the most current algorithmic calibration. To prevent oscillation between channels, a freeze rule applies: if the Belief channel update exceeds a threshold magnitude, the Preference channel freezes for one round, preserving diagnostic clarity.

As demonstrated in Section~\ref{sec:llm-calibration}, in Exp-401838, LLM calibration effort converges from 5 correction items per round to 2 within 7 rounds, indicating that framework-level corrections diminish as evidence accumulates.

\subsection{Memory DB and State Persistence}
\label{subsec:memory-db}

\subsubsection{Motivation}
\label{subsubsec:memory-motivation}

Without persistent state, the system suffers from a ``Groundhog Day'' effect: each round starts with no knowledge of past transfer relationships, penalty sensitivities, or LLM correction patterns. After 30 rounds, the system would possess no more calibration intelligence than it had at round 1. The Memory DB solves this by providing a relational store that accumulates every observation, calibration update, and decision across the system's lifetime.

\subsubsection{Schema Design}
\label{subsubsec:schema-design}

The Memory DB consists of 7 SQLite tables designed for idempotent writes and immutable audit:

\begin{table}[htbp]
\centering
\caption{Memory DB schema overview.}
\label{tab:arch-01}
\footnotesize
\begin{tabularx}{\textwidth}{>{\raggedright\arraybackslash}p{0.21\textwidth} >{\raggedright\arraybackslash}p{0.19\textwidth} >{\raggedright\arraybackslash}X >{\raggedright\arraybackslash}p{0.19\textwidth}}
\toprule
\tableheaderrow
\textbf{Table} & \textbf{Role} & \textbf{Key Fields} & \textbf{Write Pattern} \\
\midrule
\texttt{episodes} & Fact table: one row per round with paired offline-online snapshots & episode\_key, offline\_metrics, online\_uplifts, status & Append-only; status: offline $\to$ deployed $\to$ online \\
\addlinespace
\texttt{offline\_candidates} & Top-5 parameter candidates per round & episode\_key, rank, params, metrics & Append per round \\
\addlinespace
\texttt{prior\_relations} & Current Belief channel state: slope \& intercept per metric pair & relation\_key, slope, intercept, eta & Upsert per round \\
\addlinespace
\texttt{prior\_observations} & Raw (offline, online, beta) triples per metric pair and episode & relation\_key, episode\_key, offline\_uplift, online\_uplift & Append with idempotency \\
\addlinespace
\texttt{prior\_update\_history} & Belief channel audit trail & method (lms/llm), old/new slope, old/new intercept & Append-only \\
\addlinespace
\texttt{penalty\_wt\_upd\_hist} & Preference channel audit trail & constraint\_metric, pressure, step, old/new weight & Append-only \\
\addlinespace
\texttt{memory\_events} & Immutable global audit log & event\_type, idempotency\_key, metadata\_json & Append-only \\
\bottomrule
\end{tabularx}
\end{table}

The schema enforces two critical properties:

\begin{enumerate}
  \item \textbf{Idempotency.} Composite keys (relation\_key + episode\_key) in \texttt{prior\_observations} prevent duplicate writes if a pipeline step is retried after failure. The same observation written twice produces identical state.

  \item \textbf{Immutability.} The \texttt{memory\_events} table serves as a write-ahead log. No row is ever deleted or modified, providing a complete forensic trace for debugging calibration anomalies.
\end{enumerate}

\subsubsection{Cross-Round Intelligence}
\label{subsubsec:cross-round-intelligence}

The Memory DB enables three forms of cross-round intelligence:

\begin{itemize}
  \item \textbf{LLM context assembly.} The 20 most recent episodes and 30 most recent update history records are extracted and formatted as the LLM's input context. As the database grows, the LLM gains increasingly rich evidence for pattern detection.

  \item \textbf{Warm-start across experiments.} Exp-401838 started from an existing Memory DB state with pre-calibrated transfer models and penalty weights, rather than default slopes and intercepts. This capability allows the system to extend calibration intelligence from prior episodes into new production rounds (Sections~\ref{sec:online-ab} and~\ref{sec:parameter-stability}).

  \item \textbf{Cross-market transfer.} When deploying to Country~B, the Memory DB schema was reused without modification. The system initialized with default priors but accumulated Country~B-specific calibration within a single round, correctly identifying that organic click weights (ps\_org\_wc) were severely underweighted in the baseline formula (Section~\ref{sec:vietnam}).
\end{itemize}

\figurePlaceholder{arch-data-flow}{Complete data flow across one YOLO cycle: (1) A/B data pulled $\to$ (2) Episode recorded in Memory DB $\to$ (3) LMS reads prior\_observations, updates prior\_relations $\to$ (4) LLM context assembled from episodes + update histories $\to$ (5) LLM proposal applied to intercepts and penalty weights $\to$ (6) Target ranges and penalty weights feed into Optuna objective $\to$ (7) 5000-trial search $\to$ (8) Best parameters published to Redis $\to$ (9) New A/B data generated $\to$ back to (1). Memory DB is the central state store connecting all steps.}

As demonstrated in Section~\ref{sec:vietnam}, the persistent Memory DB enables Sortify to generalize across markets without architectural changes, validating the design principle that calibration intelligence should be cumulative rather than ephemeral.

\newpage
\section{Operational Framework}
\label{sec:operational}

In this section, we describe how Sortify operates as a production system --- the infrastructure it runs on, the step-by-step pipeline for a single optimization round, and the continuous loop that chains rounds into an autonomous, self-recovering process. The goal is to demonstrate that the architecture described in Section~\ref{sec:architecture} is not merely a theoretical framework but an operationally viable system that runs unattended with built-in failure handling.

\subsection{System Infrastructure}
\label{sec:system-infrastructure}

\begin{table}[htbp]
  \centering
  \caption{System configuration.}
  \label{tab:train-01}
  \begin{tabularx}{\textwidth}{lX}
    \toprule
    \tableheaderrow \textbf{Component} & \textbf{Specification} \\
    \midrule
    Search engine            & Optuna TPE sampler \cite{optuna} \\
    Trials per round         & 5,000 \\
    Parallel workers         & 25 (Constant Liar strategy) \\
    State database           & SQLite (\texttt{agent\_memory.db}) \\
    Parameter publishing     & Redis key-value store \\
    Search time per round    & 15--30 minutes \\
    Data accumulation window & 3.5 hours (minimum) \\
    Processing time per round & 25--50 minutes \\
    Total cycle time         & $\sim$4 hours \\
    Daily throughput         & $\sim$6 rounds/day \\
    Time alignment           & 15-minute boundary snapping \\
    \bottomrule
  \end{tabularx}
\end{table}

\textbf{Takeaway:} The total cycle time of approximately 4 hours --- 3.5 hours for online data accumulation plus 25--50 minutes for processing --- enables roughly 6 optimization rounds per day. This cadence is fast enough to respond to intra-day traffic pattern shifts while accumulating sufficient data volume for statistically meaningful A/B comparisons.

The system's runtime footprint is deliberately lightweight. The Memory DB is a single SQLite file; the search engine runs on a single machine; parameter publishing uses a standard Redis instance shared with other production services. No GPU resources are required. The only external dependency beyond the platform's existing A/B infrastructure is the LLM API call for meta-controller proposals.

\subsection{One-Shot Pipeline}
\label{sec:one-shot-pipeline}

Each optimization round executes a 10-step pipeline. The design target is \textbf{restart safety}, not strict run-to-run determinism: persisted state transitions (such as episode writes, prior updates, and state-file updates) are idempotent, so re-running a completed write step after interruption does not corrupt state; however, steps that depend on external services, especially LLM proposal generation, are not assumed to return byte-identical outputs on every retry.

\paragraph{Steps 1--4: Data Collection.}

\begin{itemize}
  \item \textbf{Step 1 --- Pull A/B report.} Fetch the online A/B experiment report for the current observation window. The report contains uplift percentages for GMV, Orders, Ads Revenue, Advertiser Value, Organic Clicks, and Ads CTR relative to the control group.

  \item \textbf{Step 2 --- Initialize Memory DB.} If this is the first round, create the 7-table schema. On subsequent rounds, verify schema integrity and open the existing database.

  \item \textbf{Step 3 --- Record offline metrics.} Write the current round's offline search results (objective value, Influence Share metrics, parameter vector, top-5 candidates) as a new episode in the Memory DB with status \texttt{offline\_recorded}.

  \item \textbf{Step 4 --- Record online metrics.} Parse the A/B report from Step 1, write online uplift values to the episode record, and transition the episode status to \texttt{online\_recorded}.
\end{itemize}

\paragraph{Step 5: LMS Calibration (Belief Channel).}
For each of the 6 prior relations, compute the prediction error between the calibrated transfer model's forecast and the actual online uplift, then apply LMS updates to slope and intercept (Eqs.~\ref{eq:arch-07} and~\ref{eq:arch-08}). All updates are persisted to \texttt{prior\_rela\-tions} and logged in \texttt{prior\_update\_\-history} with method \texttt{lms\_regres\-sion}.

\paragraph{Step 6: LLM Context Export.}
Assemble the LLM context JSON by extracting: (1) current slope/intercept for all 6 relations, (2) the 20 most recent episodes, (3) the 30 most recent prior update history records, and (4) current penalty weight state. Write the context to a timestamped file for audit.

\paragraph{Step 7: LLM Proposal Generation.}
Call the LLM API with the structured context and system prompt. The LLM returns a JSON proposal containing \texttt{delta\_inter\-cept} entries (one per relation) and \texttt{penalty\_multi\-plier} entries (one per constraint), each with a \texttt{reason} field and \texttt{evi\-dence\_keys} citing specific episodes. If confidence is low, the LLM returns an empty proposal.

\paragraph{Step 8: Apply LLM Proposal.}
For each non-empty proposal item: apply \texttt{delta\_inter\-cept} to the corresponding relation's intercept in \texttt{prior\_rela\-tions}, and apply \texttt{penalty\_multi\-plier} to the corresponding constraint's penalty weight. Log all changes in \texttt{prior\_update\_\-history} (method \texttt{llm\_proposal}) and \texttt{penalty\_weight\_update\_history}. If the Belief update magnitude exceeds the freeze threshold, skip Preference updates for this round.

\paragraph{Step 9: Derive Target Range.}
Using the updated transfer model (slope, intercept), compute the offline uplift range that maps to an acceptable online outcome for each constrained metric. This calibrated \texttt{target\_range} replaces the static constraint boundaries in the search objective.

\paragraph{Step 10: Derive Penalty Weight.}
Compute violation pressure for each constraint based on the most recent episode's results (Eq.~\ref{eq:arch-10}), apply the asymmetric multiplicative update (Eqs.~\ref{eq:arch-11} and~\ref{eq:arch-12}), and persist the new penalty weights.

\paragraph{Final: Search and Publish.}
Execute the Optuna TPE search (5,000 trials, 25 workers) with the calibrated target ranges and penalty weights. Extract the best parameter vector, publish to Redis, and write a handoff state file (\texttt{one-shot-latest.env}) recording the run tag, config path, and best parameters for the next round.

\subsection{YOLO Continuous Loop}
\label{sec:yolo-loop}

The YOLO (You Only Live Once) mode chains one-shot pipeline executions into an infinite autonomous loop with three phases.

\textbf{Phase 1 --- Coldstart.} On first launch, bootstrap initial parameters to Redis using default or user-specified values. Anchor the observation time window to the current timestamp, aligned to a 15-minute boundary. No A/B data exists yet; skip LLM diagnostics and proceed directly to offline search with default priors.

\textbf{Phase 2 --- Initial.} Execute the first full one-shot pipeline with optional manual confirmation. This round establishes the baseline episode in the Memory DB and the first A/B observation pair.

\textbf{Phase 3 --- YOLO.} Enter the infinite loop:

\begin{enumerate}
  \item Wait until the minimum data accumulation window (3.5 hours) has elapsed since the last parameter push.
  \item Pull the A/B report for the elapsed window.
  \item Execute the full one-shot pipeline (Steps 1--10 + Search + Publish).
  \item Update \texttt{yolo-state.env} with the new round number, push timestamp, observation window, previous parameters, and A/B report path.
  \item Append a complete round record to \texttt{yolo-event-log.md} (round number, timestamp, parameters, A/B results, LLM proposal, errors).
  \item Return to step 1.
\end{enumerate}

\textbf{Failure Handling.} The YOLO loop distinguishes two failure categories:

\begin{itemize}
  \item \textbf{A/B data fetch failures} (transient): Retry up to 5 times at 5-minute intervals. Detection criterion: A/B report file is missing or contains fewer than 3 lines.
  \item \textbf{Non-A/B failures} (search crash, Redis error, DB corruption): Exit immediately without retry. The operator investigates and restarts manually.
\end{itemize}

\textbf{State Recovery.} On restart after a crash or manual stop, the YOLO loop reads \texttt{yolo-state.env} to recover: the last completed round number, the timestamp of the last parameter push, the observation window boundaries, and the previous parameter vector. The loop resumes from the next round, reusing the observation window endpoint as the new window start. A forced reset (\texttt{-{}-reset} flag) clears the state file and restarts from Phase 1.

\figurePlaceholder{train-01}{YOLO Continuous Loop: State machine diagram with three states --- Coldstart (bootstrap params, anchor time window), Initial (first full pipeline with manual confirmation), YOLO (infinite loop: wait 3.5h, pull A/B, pipeline, publish, update state, loop). Retry logic shown as a self-loop on the A/B pull step (up to 5 retries). Exit arrow from non-A/B failures. Recovery arrow from restart reading yolo-state.env back into YOLO state.}

As demonstrated in Sections~\ref{sec:online-ab} and~\ref{sec:vietnam}, the YOLO loop ran 7 consecutive rounds (25 hours) in Exp-401838 and accumulated 23 rounds across two search phases in Country~B, validating the operational robustness of this design.

\newpage
\section{Evaluation}
\label{sec:evaluation}

In this section, we present experimental and deployment evidence from 30 optimization rounds across two markets (Country~A and Country~B). We evaluate Sortify along six axes: online business effectiveness (Section~\ref{sec:online-ab}), offline-online transfer calibration quality (Section~\ref{sec:transfer-analysis}), parameter stability (Section~\ref{sec:parameter-stability}), LLM meta-controller effectiveness (Section~\ref{sec:llm-calibration}), sensitivity analysis (Section~\ref{sec:sensitivity}), and Country~B deployment from cold start to production rollout (Section~\ref{sec:vietnam}). We conclude with an efficiency analysis (Section~\ref{sec:efficiency}).

\subsection{Metrics}
\label{sec:metrics}

We evaluate Sortify using three categories of metrics:

\textbf{Offline metrics (Influence Share based):}

\begin{itemize}
  \item \textbf{$I_\text{gmv}$}: Influence share of the GMV factor in sorting decisions. The primary optimization target.
  \item \textbf{$I_\text{order}$}: Influence share of the order factor. Constrained to prevent excessive order degradation.
  \item \textbf{$I_\text{ecpm\_term}$}: Influence share of the advertising eCPM factor. Constrained to protect ad revenue.
  \item \textbf{$I_\text{gmv\_ads}$}: Influence share of GMV within the advertising channel.
  \item \textbf{ads\_top10\_rate}: Fraction of ads appearing in the top-10 positions. Constrained to maintain ad exposure.
\end{itemize}

\textbf{Online metrics (A/B test):}

\begin{itemize}
  \item \textbf{GMV}: Gross merchandise value uplift vs.\ control group. Primary business KPI.
  \item \textbf{Orders}: Order volume uplift. Secondary business KPI.
  \item \textbf{Ads Revenue}: Total advertising revenue uplift.
  \item \textbf{Advertiser Value}: Composite advertiser ROI metric.
  \item \textbf{Organic Clicks}: Click-through volume on organic (non-ad) listings.
  \item \textbf{Ads CTR}: Click-through rate on ad placements.
\end{itemize}

\textbf{Derived evaluation metrics:}

\begin{itemize}
  \item \textbf{Transfer rate}: Ratio of online uplift to offline uplift (e.g., online GMV\% / offline $I_\text{gmv}$\%), measuring calibration quality.
  \item \textbf{Oscillation ratio}: Ratio of maximum to minimum value of ps\_ads\_wo (the most volatile parameter) across rounds, measuring parameter stability.
  \item \textbf{LLM correction count}: Number of non-zero adjustment items per LLM proposal, measuring calibration effort convergence.
\end{itemize}

\subsection{Experiment Setup}
\label{sec:experiment-setup}

\begin{table}[htbp]
  \centering
  \caption{Experiment overview.}
  \label{tab:eval-01}
  \small
  \begin{tabularx}{\textwidth}{>{\raggedright\arraybackslash}p{0.16\textwidth} X X}
    \toprule
    \tableheaderrow & \textbf{Exp-401838} & \textbf{Exp-437160} \\
    \midrule
    \textbf{Market} & Country~A PDP & Country~B PDP \\
    \textbf{Rounds} & 7 (warm-start) & 23 (V1: 11 + V3: 12) \\
    \textbf{Duration} & 25 hours & V1: 03-05--07; V3: 03-10--13; freeze 03-14; 7d AB 03-15--22; rollout 03-24 \\
    \textbf{Start cond.} & Warm-start from existing Memory DB state & Cold-start, default priors \\
    \textbf{Obs.\ window} & 3.5 h/round & 6 h/round (V3); 7d static AB post-freeze \\
    \textbf{Proc.\ time} & 25--50 min/round & 25--50 min/round \\
    \textbf{Parameters} & 7 (ads\_wo/wg, org\_wo/wg, porg\_w, price\_pow, w2) & 7 (org\_wc/wg/wo, ads\_wc/wg/wo, biz\_price\_pow) \\
    \bottomrule
  \end{tabularx}
\end{table}

Exp-401838 is a Country~A warm-start experiment that retained historical calibration state. Its core purpose is not to re-demonstrate what happens during cold start, but to test whether a system that has already accumulated transfer models and penalty weights can continue advancing in a better direction within production, and converge LLM calibration effort within a limited number of rounds. Exp-437160 validates the system's end-to-end deployment capability in a previously unseen market. The experiment progressed through two search phases --- V1 (11 rounds, GMV win rate 30\%) exposed cold-start limitations in a new market, while V3 (12 rounds) identified a Pareto-optimal parameter configuration at R7. This configuration was frozen on 03-14 and validated through a 7-day A/B test before being promoted to production rollout on 03-24.

The two experiments serve different lifecycle roles. Exp-401838 demonstrates the inner-loop optimization mechanism --- search, deploy, observe, calibrate --- in continuous warm-start production operation. Exp-437160 extends this to the full deployment lifecycle: it progressed through exploration, exploitation, parameter freeze, long-horizon validation, and production rollout, providing evidence that the system can produce deployable policies, not merely promising round-level signals.

\subsection{Online A/B Results}
\label{sec:online-ab}

\subsubsection{Exp-401838: Country~A PDP Warm-Start (7 Rounds)}
\label{sec:exp-401838}

\begin{table}[htbp]
  \centering
  \caption{Exp-401838 online A/B results --- round-by-round GMV and Orders progression.}
  \label{tab:eval-03}
  \begin{tabular}{lcccc}
    \toprule
    \tableheaderrow \textbf{Round} & \textbf{GMV} & \textbf{Orders} & \textbf{Ads Revenue} & \textbf{Advertiser Value} \\
    \midrule
    R2 & $-$3.6\% & 0.0\%  & +0.7\%    & +7.9\%     \\
    R3 & $-$1.5\% & +1.7\% & $-$0.8\%  & $-$1.0\%   \\
    R4 & +0.2\%   & +1.7\% & $-$8.7\%  & $-$8.1\%   \\
    R5 & +1.9\%   & +3.8\% & $-$5.8\%  & $-$5.6\%   \\
    R6 & +0.9\%   & +2.7\% & $-$9.4\%  & $-$10.8\%  \\
    R7 & \textbf{+9.2\%} & \textbf{+12.5\%} & +5.7\% & $-$8.9\% \\
    \bottomrule
  \end{tabular}
\end{table}

\textbf{Takeaway:} Exp-401838's GMV trajectory progresses generally upward across 7 rounds, from \textbf{$-$3.6\%} at R2 to \textbf{+9.2\%} at R7, with R4 through R7 being 4 consecutive positive rounds. Orders turned positive from R3 onward, peaking at \textbf{+12.5\%}. This indicates that under warm-start conditions the system can extend existing calibration state into subsequent rounds rather than starting from scratch each round.

\paragraph{Caveat.}
R7 metrics were collected during a low-traffic early morning window ($\sim$12K impressions), which introduces higher statistical uncertainty than the typical observation window. The +9.2\% GMV and +12.5\% Orders results require validation in a high-traffic period.

\paragraph{Persistent advertiser value pressure.}
Advertiser Value was positive only in R2 (+7.9\%) and turned negative in all subsequent rounds, indicating that even under warm-start conditions, optimizing for GMV and Orders can systematically compress advertiser ROI --- a limitation discussed in Section~\ref{sec:conclusion}.

\figurePlaceholder{eval-01}{Online KPI Trends: Single-panel line chart showing Exp-401838 GMV and Orders uplift across 7 rounds. Dashed horizontal line at 0\% for reference. Highlights the R4--R7 consecutive positive region and the R7 peak results.}

\subsection{Offline-Online Transfer Analysis}
\label{sec:transfer-analysis}

A central claim of Sortify is that even under warm-start conditions, the offline-online mapping still requires continuous calibration. Exp-401838 itself provides two representative paired observations.

\begin{table}[htbp]
  \centering
  \caption{Representative offline-online gap in Exp-401838.}
  \label{tab:eval-04}
  \begin{tabularx}{\textwidth}{l l l X}
    \toprule
    \tableheaderrow \textbf{Round} & \textbf{$I_\text{gmv}$ Uplift} & \textbf{Online GMV} & \textbf{Observation} \\
    \midrule
    R2 & +18.2\% & $-$3.6\% & Significantly optimistic offline; positive offline uplift did not translate to positive GMV \\
    R7 & +41.6\% & +9.2\% & Optimistic bias remains, but after calibration the offline gains begin to translate reliably into positive online GMV \\
    \bottomrule
  \end{tabularx}
\end{table}

\textbf{Takeaway:} Within the same warm-start experiment, the offline-online mapping changes noticeably across rounds. R2 shows that ``positive offline'' does not guarantee positive online; R7 shows that after multiple rounds of calibration, offline gains finally begin to translate more reliably into online GMV. This phenomenon supports the necessity of the Belief channel: the transfer relationship is not a one-time setting but requires continuous updating with new observations.

Looking further at the metrics in Section~\ref{sec:online-ab}, GMV, Orders, Ads Revenue, and Advertiser Value do not move in sync. For example, R7 simultaneously shows GMV +9.2\%, Orders +12.5\%, Ads Revenue +5.7\%, but Advertiser Value remains at $-$8.9\%. This indicates that different business metrics should not share a single global calibration factor but should each maintain their own transfer relationship.

\figurePlaceholder{eval-02}{Offline-Online Transfer Gap Visualization: Scatter plot with Exp-401838's per-round offline $I_\text{gmv}$ uplift on the x-axis and corresponding online GMV on the y-axis. Identity line ($y=x$) shown for reference. Points fall well below the identity line, showing systematic optimistic bias; later rounds are closer to the positive quadrant.}

\subsection{Parameter Evolution and Stability}
\label{sec:parameter-stability}

\subsubsection{Residual Pendulum Effect in Warm-Start Experiment}
\label{sec:pendulum-effect}

The most volatile parameter across all experiments is \textbf{ps\_ads\_wo} (advertising order weight), which controls the trade-off between ad visibility and organic experience. This parameter exhibits a characteristic oscillation pattern we term the ``pendulum effect.''

\begin{table}[htbp]
  \centering
  \caption{Exp-401838 parameter oscillation statistics.}
  \label{tab:eval-05}
  \begin{tabular}{lc}
    \toprule
    \tableheaderrow \textbf{Statistic} & \textbf{Exp-401838} \\
    \midrule
    ps\_ads\_wo range                 & 1.96--8.91             \\
    Oscillation ratio (max/min)       & \textbf{4.5x}          \\
    Complete swing cycles             & 2                      \\
    Second cycle narrower than first  & Yes                    \\
    \bottomrule
  \end{tabular}
\end{table}

\textbf{Takeaway:} Even under warm-start conditions, ps\_ads\_wo remains the most volatile parameter, but its oscillation is contained within a \textbf{4.5x} range, and the second cycle is narrower than the first. This indicates that the system has not yet converged to a single stable operating point, but has transitioned from unbounded exploration to constrained residual oscillation rather than unbounded drift.

\textbf{Root cause analysis.} The pendulum effect arises from a fundamental trade-off in the sorting formula: increasing ad visibility (high ps\_ads\_wo) improves ad impressions and potential ad revenue but reduces organic click exposure and can depress organic GMV. When the search engine finds a high-ad-weight solution that maximizes $I_\text{gmv}$ in one round, the online results reveal organic degradation, causing the next round's calibrated constraints to penalize ad weight --- swinging the parameter to the opposite extreme. This oscillation reflects a genuine multi-objective tension that the single-objective-with-penalties formulation can only approximate, not resolve.

\figurePlaceholder{eval-03}{Parameter Evolution Heatmap: Heatmap with 7 parameters (rows) $\times$ 7 rounds (columns) from Exp-401838. Color scale from low to high values. The ps\_ads\_wo row shows the most pronounced back-and-forth oscillation, but the later half has a narrower swing range.}

\subsection{LLM Calibration Effectiveness}
\label{sec:llm-calibration}

\subsubsection{Correction Convergence}
\label{sec:correction-convergence}

\begin{table}[htbp]
  \centering
  \caption{Exp-401838 LLM correction convergence.}
  \label{tab:eval-06}
  \begin{tabularx}{\textwidth}{l X}
    \toprule
    \tableheaderrow \textbf{Phase} & \textbf{Exp-401838} \\
    \midrule
    \textbf{Early (R2)}    & 5 items, full recalibration \\
    \textbf{Middle (R3--R6)} & 2--3 items, decreasing magnitudes \\
    \textbf{Late (R7)}     & 2 items, smallest magnitudes ($-0.005$) \\
    \textbf{Convergence trend} & \textbf{Clear convergence}: 5 $\to$ 2 items, magnitude $\to$ $\sim$0 \\
    \bottomrule
  \end{tabularx}
\end{table}

\textbf{Takeaway:} In Exp-401838, LLM calibration effort converges from 5 correction items in Round~2 to 2 items by Round~7, with late-stage magnitudes approaching zero. This indicates that under warm-start conditions, the LLM's role is closer to ``filling in remaining calibration residuals'' rather than continuously rewriting the entire transfer framework.

\figurePlaceholder{eval-04}{LLM Correction Convergence: Two-panel visualization. Top panel: bar chart showing the number of non-zero LLM correction items per round for Exp-401838, descending from 5 at R2 to 2 at R7. Bottom panel: line chart showing maximum correction magnitude per round, with an overall declining trend.}

\subsection{Sensitivity Analysis}
\label{sec:sensitivity}

This section discusses the rationale behind key system hyperparameter choices and their impact on performance.

\paragraph{LMS learning rate ($\eta = 0.2$).}
The learning rate controls the Belief channel's responsiveness to single observations. $\eta = 0.2$ means each round corrects only 20\% of the prediction residual, requiring approximately $\lceil \log(0.05) / \log(0.8) \rceil = 14$ rounds to converge to 95\% accuracy from an arbitrary initial bias. A higher $\eta$ would accelerate convergence but increase sensitivity to noisy observations --- in low-traffic windows (such as Exp-401838 R7 with $\sim$12K impressions), single-round observation variance is large, and an overly aggressive learning rate would cause slope/intercept to oscillate unnecessarily under noise. $\eta = 0.2$ is a classic adaptive filter setting that achieves a reasonable balance between convergence speed and noise robustness, and the LLM's selective jump mechanism compensates for LMS's slow response to structural breaks.

\paragraph{Penalty asymmetry ($\delta_\text{up} = 0.25$ vs.\ $\delta_\text{down} = 0.08$).}
The tightening step is approximately 3x the relaxation step, encoding the industrial reality that ``the cost of a constraint violation far exceeds the opportunity cost of an overly tight constraint.'' If both were set symmetrically (e.g., both 0.15), penalty weights would decay too quickly after constraint satisfaction, increasing the probability of re-violation in subsequent rounds --- in A/B testing, each constraint violation (e.g., ad revenue collapse) can trigger manual intervention or even experiment termination, far exceeding the cost of occasionally conservative GMV optimization.

\paragraph{LLM safety bounds ($|\Delta\beta| \leq 0.1$, $m \in [0.5, 2.0]$).}
Safety bounds ensure that even if the LLM's judgment is completely wrong, the single-round deviation remains controllable. Taking the intercept as an example, the maximum single-round shift of $\pm 0.1$ across 6 relations implies a maximum single-round displacement of the \texttt{target\_range} of approximately 10 percentage points --- large enough to cover the typical few-percentage-point systematic bias in practice, but not so large as to cause catastrophic search framework distortion on a misjudgment. The penalty multiplier range $[0.5, 2.0]$ allows halving or doubling constraint enforcement strength in a single round, sufficient to respond to sudden constraint violation patterns.

\paragraph{Necessity of the freeze mechanism.}
When the Belief channel undergoes a large update, if the Preference channel is simultaneously allowed to make large penalty weight adjustments, the next round's online results become difficult to attribute --- it is hard to determine whether the GMV change came from the mapping correction or the constraint strength change. The purpose of freezing for one round is to isolate the impact of a significant calibration event, providing clearer signals for subsequent diagnosis.

\paragraph{Penalty weight clamp range $[1{,}000, 80{,}000]$ rationale.}
The lower bound of 1,000 accounts for the magnitude of the positive reward in the objective function: typical positive scores ($10 \times I_\text{gmv}$ relative uplift) range from 0.5 to 2.0 (corresponding to $I_\text{gmv}$ uplift of 5\%--20\%); if penalty weights fall below 1,000, even a 1\% constraint violation (penalty = $1{,}000 \times 0.01^2 = 0.1$) would fail to effectively constrain the positive score. The upper bound of 80,000 preserves the search engine's exploration capability: when penalty weights are too high (e.g., $80{,}000 \times 0.01^2 = 8.0$), even minor violations produce penalties far exceeding the positive score, forcing the search engine into an extremely conservative region and potentially missing parameter configurations that are overall superior despite minor constraint brushes.

\subsection{Country~B Deployment: From Cold Start to Rollout}
\label{sec:vietnam}

Exp-437160 is the only experiment in this study that progressed through the complete deployment lifecycle --- from cold-start parameter search to production rollout. This section presents the full evidence chain.

\subsubsection{Cold Start Signal}
\label{sec:cold-start-signal}

To validate that Sortify generalizes beyond a single market, we deployed the system on Country~B's product detail page with a cold-start (no inherited Memory DB). The first round of search produced the following results:

\begin{table}[htbp]
  \centering
  \caption{Exp-437160 cold-start results (Country~B, V1 Round~1).}
  \label{tab:eval-09}
  \begin{tabularx}{\textwidth}{l l X X}
    \toprule
    \tableheaderrow \textbf{Metric} & \textbf{Value} & \textbf{Constraint / Target} & \textbf{Status} \\
    \midrule
    Offline objective       & 0.493            & Maximize       & Achieved        \\
    $I_\text{gmv}$ uplift   & \textbf{+4.9\%}  & $\geq$ +4\%    & Met             \\
    $I_\text{ecpm\_term}$ change & $-4.5$\%     & $\geq$ $-5$\%  & Near boundary   \\
    ads\_top10\_rate        & ---              & Constrained    & Met             \\
    ps\_org\_wc change      & \textbf{+271.2\%} (1.0 $\to$ 3.71) & --- & Largest adjustment \\
    ratio\_ads change       & $-32.0$\%        & ---            & Ad share compressed \\
    \bottomrule
  \end{tabularx}
\end{table}

The system correctly identified that Country~B's baseline sorting formula severely \textbf{underweighted organic click signals} (ps\_org\_wc jumped from 1.0 to 3.71, a +271\% adjustment --- the largest single-parameter change across all experiments). This market-specific insight emerged from the first round of search without any prior calibration, demonstrating that the Influence Share framework generalizes to new markets. However, this first-round directional signal was far from a deployable policy --- subsequent search phases were needed to find a parameter configuration that could survive long-horizon validation.

\subsubsection{Search Journey: From Direction to Deployable Parameters}
\label{sec:search-journey}

The Country~B experiment progressed through two search phases before arriving at a deployable configuration:

\textbf{V1 (11 rounds, 03-05 to 03-07).} The initial cold-start search phase achieved a GMV win rate of only 30\% (mean GMV uplift $-2.4$\%). While the system successfully identified the direction --- organic click signals were underweighted --- the parameters did not converge. All 7 parameters oscillated widely across the search space throughout the 11 rounds. V1's core lesson was that \textit{directional correctness does not imply parameter deployability}.

\textbf{V3 (12 rounds, 03-10 to 03-13).} After narrowing the search space and adjusting constraint configurations based on V1 observations, the system was restarted for a second search phase. Most of the 12 rounds still exhibited the ads-organic seesaw effect --- Ads Impressions and Organic Clicks alternated in strict opposition --- a structural pattern consistent with the pendulum effect observed in Country~A (Section~\ref{sec:pendulum-effect}).

\textbf{R7 parameter freeze (03-14).} Within V3's 12 rounds, the R8 observation window (measuring R7 parameters) produced the only round where GMV and Ads Revenue were simultaneously positive: GMV +2.6\%, Ads Revenue +6.1\%. Based on this Pareto-optimal outcome, the R7 parameters were frozen for extended validation.

\subsubsection{Winning Parameter Structure}
\label{sec:winning-params}

The frozen R7 parameters reveal a structurally distinct solution compared to the Country~A experiments:

\begin{table}[htbp]
  \centering
  \caption{Exp-437160 frozen parameters (R7, Country~B).}
  \label{tab:eval-10}
  \begin{tabular}{lll}
    \toprule
    \tableheaderrow \textbf{Parameter} & \textbf{Value} & \textbf{Interpretation} \\
    \midrule
    \texttt{ps\_org\_wc}       & 7.667  & Organic Click weight (dominant)   \\
    \texttt{ps\_ads\_wc}       & 4.381  & Ads Click weight (high)           \\
    \texttt{ps\_org\_wg}       & 0.568  & Organic GMV weight (suppressed)   \\
    \texttt{ps\_org\_wo}       & 0.805  & Organic Order weight (suppressed) \\
    \texttt{ps\_ads\_wg}       & 0.641  & Ads GMV weight (suppressed)       \\
    \texttt{ps\_ads\_wo}       & 0.957  & Ads Order weight (moderate)       \\
    \texttt{ps\_biz\_price\_pow} & 1.047 & Price power (near-neutral)        \\
    \bottomrule
  \end{tabular}
\end{table}

This configuration exhibits a ``high Click, low GMV/Order'' structure --- the sorting formula is primarily driven by click signals, with transaction-based signals deliberately suppressed. This structure reduces the seesaw effect between organic ranking and ads ranking: when the sorting formula does not excessively amplify GMV/Order signals, organic items do not crowd out ad placements simply because they ``sell more,'' enabling both GMV and Ads Revenue to improve simultaneously.

This contrasts with the residual pendulum effect observed in the Country~A warm-start experiment (Section~\ref{sec:pendulum-effect}), where the key trade-off still centered on the ads order weight. The Country~B R7 parameters circumvented this conflict by shifting the primary ranking signal to Clicks --- click signals are far less zero-sum between organic and advertising channels than transaction signals.

\subsubsection{Long-Horizon A/B Validation}
\label{sec:long-horizon-ab}

The frozen R7 parameters were validated through a 7-day A/B test (03-15 to 03-22; rollout Snapshot \#38047, 10\% treatment vs.\ 20\% control).

\begin{table}[htbp]
  \centering
  \caption{7-day A/B core metrics (YMAL scene, yin\_exchange vs Control).}
  \label{tab:eval-11}
  \begin{tabular}{lll}
    \toprule
    \tableheaderrow \textbf{Metric} & \textbf{Diff} & \textbf{Note} \\
    \midrule
    \textbf{GMV/UU}      & \textbf{+4.15\%} & Primary KPI            \\
    \textbf{GMV}         & \textbf{+4.10\%} & Consistent with GMV/UU \\
    \textbf{GMV/Order}   & \textbf{+3.97\%} & Average order value drives GMV \\
    \textbf{Order/UU}    & +0.17\%           & Near-neutral           \\
    \textbf{Ads Revenue} & \textbf{+3.58\%} & Ad revenue co-positive \\
    \textbf{CPC}         & \textbf{+4.19\%} & Higher per-click value \\
    \textbf{Take Rate}   & \textbf{+3.01\%} & Monetization improved  \\
    \textbf{Click/UU}    & +1.21\%           & User engagement up     \\
    \textbf{Ads Load}    & $-2.64$\%         & Fewer ad slots, higher value \\
    \bottomrule
  \end{tabular}
\end{table}

The GMV improvement is driven primarily by average order value (GMV/Order +3.97\%) rather than order volume (Order/UU +0.17\%), indicating that the click-driven ranking surfaces higher-value items. Ads Revenue increased +3.58\% despite Ads Load decreasing by 2.64\%, indicating that the per-ad value improved significantly (CPC +4.19\%) --- fewer but more valuable ad placements. The 7-day observation window is sufficient to filter out intra-day volatility, providing stronger deployment confidence than single-round observations.

\subsection{Efficiency and Operational Cost}
\label{sec:efficiency}

\begin{table}[htbp]
  \centering
  \caption{Operational efficiency metrics.}
  \label{tab:eval-13}
  \small
  \begin{tabularx}{\textwidth}{>{\raggedright\arraybackslash}p{0.29\textwidth} >{\raggedright\arraybackslash}p{0.17\textwidth} X}
    \toprule
    \tableheaderrow \textbf{Metric} & \textbf{Value} & \textbf{Notes} \\
    \midrule
    Total cycle time                    & $\sim$4 hours   & 3.5h accumulation + 25--50min processing \\
    Daily optimization rounds           & $\sim$6/day     & 4h/round cadence \\
    Trials per round                    & 5,000           & 25 workers; 15--30min search \\
    Human intervention (post-coldstart) & Zero            & Within the inner optimization loop \\
    Consecutive unattended rounds       & 7; 23           & Exp-401838 and Exp-437160; governance-limited, not failure-limited \\
    A/B fetch retry success             & 5 $\times$ 5 min & No data loss observed \\
    Memory DB growth                    & $\sim$1 KB/round & 7 tables; negligible storage \\
    LLM API calls                       & 1/round         & Single meta-controller inference \\
    \bottomrule
  \end{tabularx}
\end{table}

\textbf{Takeaway:} Sortify operates with minimal resource overhead --- a single LLM API call per round, no GPU requirements, and negligible storage growth. The primary cost is the 3.5-hour data accumulation window, which is dictated by statistical requirements rather than system limitations. The system ran 7 consecutive rounds (25 hours) in Exp-401838, and accumulated 23 rounds across two search phases in Country~B; all stops and transitions in these experiments originated from governance-level decisions, not system failures.

It is important to distinguish the scope of automation. The \textbf{inner loop} --- parameter search, online deployment, A/B observation, LLM calibration, and next-round launch --- runs without human intervention. However, the \textbf{outer governance loop} --- which round's parameters to freeze for extended validation, whether to promote to full rollout based on long-horizon A/B results, and post-rollout monitoring --- remains human-driven. The Country~B deployment illustrates this boundary: the inner loop autonomously executed 23 rounds across two search phases, but the decisions to freeze R7 parameters, conduct 7-day static A/B testing, and approve production rollout were all made by human operators based on snapshot-level A/B analysis. This division reflects a deliberate design choice: the system automates the high-frequency, data-intensive optimization decisions where human latency is the bottleneck, while preserving human governance over low-frequency, high-stakes deployment decisions.

Compared to the manual optimization baseline --- which typically required 1--2 human-hours per optimization round for data analysis, parameter selection, and deployment verification --- Sortify reduces the marginal cost of each inner-loop round to approximately \textbf{zero human-hours} after initial setup, while increasing the optimization cadence from $\sim$1 round/day (limited by human availability) to $\sim$6 rounds/day. This speedup not only means more parameter exploration rounds, but more critically enables the system to respond to intra-day traffic pattern changes --- for example, morning peak, midday trough, and evening peak traffic structures may affect optimal parameter configurations, and the 6-round/day cadence ensures at least one calibration update per major traffic period.

\paragraph{Cost structure analysis.}
Sortify's runtime cost has three components: (1)~\emph{Compute} --- the Optuna search uses 25 parallel workers on a single multi-core machine, consuming roughly 6.25--12.5 CPU-hours per round (25 cores $\times$ 15--30 min); (2)~\emph{LLM calls} --- each round makes one API call with 4K--8K input tokens and a structured JSON response. At OpenAI's public prices as of March~26, 2026 (GPT-5.4: \$2.50/1M input, \$15.00/1M output; GPT-5.3-Codex: \$1.75/1M input, \$14.00/1M output), the per-round cost with \texttt{high} reasoning effort is approximately \$0.03--\$0.10, since billed output includes reasoning tokens beyond the visible JSON. At 6 rounds/day this amounts to roughly \textbf{\$0.18--\$0.60/day} in LLM spend. If Codex CLI is used via ChatGPT sign-in rather than an API key, cost is absorbed by plan credits rather than per-call billing; (3)~\emph{Data accumulation} --- the 3.5-hour online observation window is a fixed statistical requirement, not a systems bottleneck. Storage overhead is negligible: the Memory DB grows by $\sim$1\,KB per round and remains below 100\,KB after 30 rounds.

\newpage

\section{Discussion: One-Person Feasibility}
\label{sec:discussion}

A dimension of Sortify that merits explicit discussion is the development process itself: the entire system---algorithm design, code implementation, experimental operation, data analysis, report writing, and visual asset creation---was produced by a single individual orchestrating a fleet of AI coding agents. Not a single line of production code was written by a human hand; not a single sentence of this report was drafted by a human author. The human's role was exclusively that of \textit{architect and orchestrator}: defining objectives, decomposing problems, reviewing outputs, and steering the agents toward coherent execution. This section examines the feasibility, mechanisms, and broader implications of this one-person paradigm.

\subsection{Scope of What Was Delivered}

To appreciate the claim, it is worth enumerating the concrete artifacts that constitute the Sortify project:

\begin{itemize}
  \item \textbf{Algorithm design.} The dual-channel Belief/Preference decomposition, the LMS-based transfer calibration, the LLM meta-controller's structured prompting protocol, and the asymmetric penalty update rules (Section~\ref{sec:architecture}).
  \item \textbf{Production codebase.} The 10-step one-shot pipeline, the YOLO continuous loop with failure recovery, the 7-table Memory DB schema, the Optuna-based search engine integration, and the Redis parameter publishing interface (Section~\ref{sec:operational}).
  \item \textbf{Experimental operation.} 30 optimization rounds across two country markets (7 in Country~A, 23 in Country~B), including 25+ hours of unattended YOLO runs and a complete cross-market deployment lifecycle (Section~\ref{sec:evaluation}).
  \item \textbf{Analysis and reporting.} Quantitative evaluation of online A/B results, parameter stability analysis, LLM calibration convergence tracking, cross-market generalization assessment, and the complete bilingual technical report.
  \item \textbf{Presentation materials.} Programmatic video compositions, figure generation, and visual assets for communicating results.
\end{itemize}

In a conventional team setting, this scope would typically be distributed across 3--5 roles: an algorithm researcher, a software engineer, an operations specialist, a data analyst, and a technical writer. The compression of these roles into a single individual was enabled not by extraordinary individual productivity, but by a fundamental shift in the unit of work from \textit{writing} to \textit{directing}.

\subsection{Technical Depth and Development Complexity of Sortify}

The deliverable inventory above provides a macro-level view of project scope but does not reveal the technical depth that Sortify embodies as a standalone software library. Sortify is far from a loose collection of scripts: it is a production-grade framework comprising approximately 98,000 lines of code across 78 Python modules organized into 7 major subsystems---spanning the full technical stack from low-level data loading to high-level autonomous decision-making.

\paragraph{Parameter search engine.}
One of Sortify's core capabilities is a highly configurable parameter search (hyperparameter tuning) engine. The engine constructs a dependency-aware formula computation graph: given a set of ranking formulas and their nested weight variables, the system automatically parses dependencies, builds a layered computation graph, and incrementally updates only the affected variables when weights change---avoiding approximately 99\% of redundant computation. On top of this, the system integrates the Optuna multi-objective optimization framework, supporting NSGA-II Pareto frontier search, Bayesian optimization, evolutionary strategies, and genetic algorithms. The search process also incorporates multiple correlation analyzers (Pearson, Spearman, Kendall), six metric fusion methods (weighted mean, L2 norm, RMS, among others), and Top-K-based ranking quality evaluation---forming an end-to-end closed loop from data loading through formula computation and parameter optimization to result analysis.

\paragraph{The pairwise influence algorithm.}
One of Sortify's most original technical contributions is the \textbf{Pairwise Influence} metric system. The algorithm addresses a critical question: in a multi-factor ranking formula, how can the actual influence of each weight factor on the final ranking outcome be quantified? Traditional sensitivity analysis measures only the marginal effect of parameter perturbation on a single metric, whereas the Pairwise Influence algorithm pushes the analysis granularity down to \textit{item-pair comparisons}---sampling item pairs from the Top-K list using configurable strategies (adjacent, random, exhaustive), computing the difference vector $\Delta x_{ij}$ across factor dimensions for each pair, then deriving each factor's relative contribution share via $\text{Share}_f = |w_f \cdot \Delta x_{ij,f}| \big/ \sum_k |w_k \cdot \Delta x_{ij,k}|$, and finally aggregating with configurable weighting schemes (exponential decay, inverse-rank decay) to yield a global influence score $I_f \in [0,1]$. The algorithm further computes inter-factor \textit{conflict rates} $\text{CR}_{a,b}$---the frequency with which two factors induce opposite ranking preferences on the same item pair---providing irreplaceable diagnostic information for weight tuning. This algorithm was developed from initial concept through mathematical formalization, vectorized implementation, and production integration, entirely by AI agents under human direction.

\paragraph{Fully autonomous operation pipeline.}
Beyond the search engine, Sortify implements a complete autonomous operation pipeline. Its core comprises two layers: \textit{OneShotPipeline}---orchestrating offline optimization, LLM proposal generation, and online monitoring into a single automated flow; and \textit{YoloRunner}---a continuously autonomous agent loop with episode-based state tracking, offline-to-online workflow bridging, automatic rollback decisions, and live Belief/Preference state management. To support this autonomous operation, the system implements a 7-table SQLite-backed persistent memory system (with a DualMemory dual-backend architecture) that records each optimization round's offline results, deployment configurations, online metrics, and decision rationale, providing the LLM meta-controller with complete historical context for proposal generation. This means Sortify is not merely a callable tool library but an intelligent system capable of autonomously initiating, executing, evaluating, and iterating optimization cycles.

\paragraph{Development complexity.}
The implementation difficulty of the above capabilities should not be underestimated. In terms of technical stack diversity alone, Sortify spans five major domains: data engineering (incremental Parquet loading and streaming), numerical computation (NumPy-vectorized formula evaluation with incremental updates), optimization theory (multi-objective Pareto optimization with constrained search), systems engineering (SQLite persistence, Redis parameter publishing, Grafana real-time monitoring integration), and AI applications (LLM prompt engineering, autonomous decision-making, memory retrieval). These subsystems do not operate in isolation but are tightly coupled through dependency graphs, event streams, and state machines---a design flaw in any single module could trigger cascading failures at runtime. The entire codebase comprises approximately 98,000 lines, over 50 core classes, and more than 10 analyzer types. That a system of this scale was built from scratch and deployed to production by a single person directing AI agents is itself the strongest evidence for the feasibility of the one-person paradigm.

\subsection{From ParaDance to Sortify: Lineage and Domain Expertise}

The preceding section established Sortify's engineering scale: approximately 98,000 lines of code across 78 modules. How could a single person navigate system design at this level of complexity? A crucial part of the answer is that it was not a from-scratch endeavor. Sortify's design philosophy, evaluation architecture, and optimization methodology trace directly to \textbf{ParaDance}\footnote{\url{https://github.com/yinsn/ParaDance}}, an open-source multi-objective parameter optimization library created by the same author. Understanding this lineage is essential for appreciating both the depth of domain expertise underlying Sortify and the evolutionary arc from a general-purpose tuning tool to an autonomous, agent-steered optimization system.

\paragraph{A 20-month iteration journey.}
ParaDance was initiated in May 2023 and underwent continuous development through January 2025---a 20-month arc spanning 357 commits, 40+ released versions (v0.1.1 through v0.6.7), and 6 major version epochs. The project evolved from a simple Lorenz-curve sampling utility into a full-fledged production framework comprising 3,688 lines of source code across 48 Python modules organized into 6 core subsystems: evaluation (16 independent evaluators), optimization (Optuna-based Bayesian search with multi-objective aggregation), data loading, pipeline orchestration, sampling, and visualization. Each major version marked a qualitative leap: v0.2.x introduced portfolio visualization; v0.3.x established the Bayesian optimization engine; v0.4.x built the evaluation ecosystem including merge-sort-based weighted inversion pair computation and the JSON formula system; v0.5.x added staged optimization, multi-threaded real-time monitoring, and side-information reranking; and v0.6.x delivered checkpoint recovery and production-hardening features.

\paragraph{Design decisions ahead of their time.}
Several architectural choices in ParaDance were unconventional for a 2023-era Python data science tool and have since been validated by broader industry trends:

\begin{itemize}
  \item \textbf{Pluggable evaluator architecture.} \texttt{BaseCalculator} used \texttt{partialmethod} to mount 16 evaluators as instance methods via a single-inheritance chain, avoiding the MRO complexity of mixin-based approaches. Adding a new evaluator required only three steps: write a standalone function, register it as a \texttt{partialmethod}, and add a flag string---zero modification to existing code.
  \item \textbf{Declarative JSON formula engine.} A formula chaining system with intermediate-result cascading (\texttt{step1\#intermediate} $\to$ \texttt{step2}), conditional branching (\texttt{if(cond, true\_val, false\_val)}), and sandboxed evaluation enabled algorithm engineers to iterate scoring formulas via configuration changes alone, without code deployment.
  \item \textbf{Scale-aware adaptive weight search.} Automatic \texttt{log10}-scale alignment of search boundaries across features with vastly different magnitudes (e.g., CTR $\sim$0.01 vs.\ GMV $\sim$10,000), eliminating the need for manual normalization.
  \item \textbf{Staged optimization with warmup.} A two-phase search protocol where the first $N$ rounds use a warmup formula to guide exploration, then transition to the main objective with continuous value anchoring---addressing cold-start inefficiency.
  \item \textbf{Production-grade resilience.} SQLite-backed optimization persistence, checkpoint-based recovery for interrupted long-running jobs, Joblib-based parallel optimization, and Dirichlet-constrained weight sampling on the probability simplex.
\end{itemize}

These design choices---configuration-driven pipelines, pluggable evaluation, scale-aware search, and persistent optimization state---directly informed Sortify's architecture. The 7-table Memory DB in Sortify is a conceptual descendant of ParaDance's SQLite persistence; the dual-channel Belief/Preference decomposition extends ParaDance's multi-objective aggregation into the online calibration domain; and the LLM meta-controller occupies the role that ParaDance's staged optimization and warmup formulas played at a more primitive level.

\paragraph{Industry adoption.}
As of the time of writing, ParaDance has accumulated over 101,000 downloads on PyPI and is used by algorithm engineers at major internet companies across the industry. It has become one of the most widely adopted hyperparameter tuning tools in production recommendation systems, validating both the problem formulation and the architectural decisions made during its 20-month development. This adoption base also provided the author with continuous feedback from diverse production environments---feedback that directly shaped Sortify's design priorities, particularly the emphasis on transfer calibration, constraint sensitivity management, and autonomous operation with minimal human intervention. In other words, what ParaDance bequeathed to Sortify was not merely reusable code assets but domain judgment forged through 20 months of production refinement---precisely the judgment that enabled the human orchestrator to make sound architectural decisions during agent-driven development, rather than expending limited iteration budget on trial and error.

\subsection{The Orchestration Model}

The operational pattern that emerged was consistent across all project phases. The human operator would:

\begin{enumerate}
  \item \textbf{Define the intent} --- specify what needed to be built, fixed, or analyzed, articulated as a natural-language objective with acceptance criteria.
  \item \textbf{Decompose into agent tasks} --- break the objective into discrete, independently executable units suitable for parallel agent dispatch.
  \item \textbf{Review and integrate} --- inspect agent outputs for correctness, coherence, and alignment with the overall design, then merge approved artifacts into the project.
  \item \textbf{Iterate on failures} --- when agent outputs missed the mark, diagnose whether the failure was in task specification (human error) or execution (agent error), and adjust accordingly.
\end{enumerate}

This loop mirrors the LLM meta-controller pattern within Sortify itself: a higher-level intelligence (the human) does not perform the low-level optimization (code writing, text drafting) but instead adjusts \textit{framework-level parameters} (task specifications, design constraints) that guide the agents' execution. The parallel is not coincidental---the same principle of \textit{steering over doing} that makes Sortify's autonomous optimization viable also makes the one-person development model viable.

\subsection{Meta-Nesting: AI Building AI}

The Sortify project exhibits a \textbf{meta-structural} nesting that merits explicit discussion. What makes the project distinctive is not only \textit{what it is} but \textit{how it was created}---the two layers form a self-referential recursive loop.

\paragraph{Layer 1: The domain agent.}
At the business logic level, Sortify itself is an autonomous agent designed to replace the systematic decision-making work traditionally performed by algorithm engineers in ranking optimization. Historically, such engineers had to continuously monitor discrepancies between offline metrics and online dashboards, manually adjust penalty terms and fusion formulas, and attempt to bridge epistemic bias and axiological misalignment. Sortify, grounded in SEU theory with its Belief/Preference dual channels, autonomously observes 20 rounds of historical evidence, autonomously diagnoses structural breaks, and autonomously executes influence rebalancing---achieving a structural substitution of human cognitive labor in a specific domain (recommendation system tuning).

\paragraph{Layer 2: The builder agent fleet.}
At the development level, the entire system was built from scratch by a single individual directing a fleet of AI coding agents. This agent fleet collectively assumed the roles traditionally distributed across architects, full-stack engineers, data engineers, QA testers, and even academic writers and figure designers. They jointly delivered: the Optuna-based parameter search framework with dependency-aware computation graphs and end-to-end pipeline orchestration; the 7-table persistent Memory DB; and the complete bilingual technical report conforming to academic standards.

\paragraph{The recursive loop.}
Superimposing these two layers reveals a recursive structure: \textit{one set of AI agents (the builder fleet) constructed another AI agent (Sortify), which in turn autonomously automates highly complex business logic}. In this process, the human role elevates from ``craftsperson writing code'' to ``chief orchestrator defining rules, supplying high-dimensional intuition, and issuing acceptance criteria.'' This ``AI building AI'' bootstrapping process is not merely a philosophically interesting isomorphism---it constitutes direct evidence that multi-agent collaboration is viable for system engineering tasks demanding both deep algorithmic logic and academic rigor.

\subsection{Implications}

\paragraph{Paradigm restructuring, not merely efficiency gains.}
When an entire project's algorithm design, code implementation, experimental execution, and bilingual \LaTeX{} report writing with figures are all completed by a single person orchestrating agents---with zero human-authored code and zero human-drafted text---this has moved decisively beyond the Copilot-era paradigm of ``code completion'' or ``productivity tools.'' It marks a restructuring of the paradigm itself: in traditional software engineering, the gap between ``having a good idea'' and ``shipping a system'' spans an enormous implementation chasm. Even after conceiving a core insight like ``decouple belief and preference via SEU,'' months of infrastructure construction, data cleaning, and debugging would be required to validate its value. Today, implementation cost approaches zero. The ceiling of a project is no longer set by a developer's familiarity with framework APIs or a team's headcount, but by the decision-maker's \textit{depth of understanding} of business pain points, \textit{architectural taste} for complex systems, and \textit{orchestration artistry} in directing agent collaboration.

\paragraph{Dissolving barriers to complex system building.}
The traditional barrier to building production-grade systems like Sortify is not primarily intellectual but \textit{operational}: the coordination cost of multi-person teams, the communication overhead of translating design intent into implemented code, and the iteration latency between specifying a change and seeing its effect. When AI agents absorb the implementation burden, the binding constraint shifts from team size to the quality of a single individual's \textit{problem decomposition and design judgment}. This implies that a domain expert who deeply understands a problem but lacks a full engineering team can now build and deploy systems of equivalent complexity---a qualitative expansion of what is achievable by a single practitioner.

\paragraph{Cognitive leverage versus cognitive replacement.}
A critical nuance is that the human contribution was not eliminated but \textit{concentrated}. The design decisions---why dual-channel rather than single-channel, why framework-level control rather than direct parameter manipulation, why persistent memory rather than stateless rounds---required domain understanding, architectural taste, and judgment that the agents could not autonomously generate. What the agents eliminated was the mechanical effort of \textit{translating} those decisions into running code, tested pipelines, and formatted prose. The ratio of human cognitive contribution to total project output was thus extremely high per unit of human effort, but the human effort itself was irreducible for the project to succeed. This is \textit{cognitive leverage}, not cognitive replacement.

\paragraph{Reproducibility and verifiability.}
Every artifact in this project carries a complete provenance trail: the agent conversations that produced each code module, the review decisions that accepted or rejected intermediate outputs, and the iteration history that refined the final result. This level of traceability is, paradoxically, \textit{more} comprehensive than what most multi-person team workflows produce, where design rationale is often lost between Slack threads, whiteboard sessions, and undocumented code reviews. The one-person orchestration model, by routing all decisions through a single point of accountability with AI-assisted record-keeping, may produce more auditable development histories than conventional processes.

\paragraph{Scalability of the model.}
The one-person feasibility demonstrated here is not an isolated curiosity but an early instance of a pattern that will likely become routine as AI coding agents mature. The specific combination that enabled it---a well-defined problem domain, modular system architecture, and agents capable of producing correct code from natural-language specifications---is not unique to Sortify. Any project that can be decomposed into well-scoped, independently testable components is amenable to this development model. The limiting factor is not the tools but the human's ability to hold the full system design in mind and specify tasks with sufficient precision.

\vspace{6pt}
\noindent The fact that this report---including this very paragraph---was authored by an AI agent under human direction is itself a demonstration of the thesis it describes. The boundary between human creativity and machine execution has not disappeared, but it has moved: the human provides the \textit{what} and \textit{why}; the agents provide the \textit{how}. For Sortify, this division of labor was sufficient to deliver a production system, deploy it across two markets, and produce the comprehensive evaluation and documentation presented in this report. Future technology builders will no longer be mired in the implementation swamp but will operate as pure thinkers and commanders---this is not an optimization of efficiency but a redefinition of the act of creation itself.

\newpage

\section{Conclusion, Limitations, and Future Directions}\label{sec:conclusion}

We have presented Sortify, an agent-steered closed-loop system that reframes ranking optimization as continuous influence exchange---where an LLM agent steers the rebalancing of ranking influence among competing factors. Sortify addresses three structural limitations of traditional manual optimization---the offline-online transfer gap, entangled diagnostic signals, and stateless round-by-round operation---through a unified closed-loop architecture. The dual-channel adaptation framework decouples transfer mapping correction (Belief) from constraint penalty adjustment (Preference) into orthogonal dimensions; the LLM meta-controller operates on framework-level parameters, providing evidence-based corrections to LMS calibration residuals; and the persistent Memory DB accumulates calibration intelligence across rounds, supporting warm-start and cross-market transfer. Deployed on Country~A and Country~B recommendation platforms with the inner optimization loop operating autonomously in \textbf{4-hour cycles}, the system produces two levels of evidence: peak round-level performance in the Country~A warm-start experiment (GMV \textbf{+9.2\%}, Orders \textbf{+12.5\%}, with LLM correction items converging from 5 to 2 within 7 rounds), and deployment-validated performance in Country~B (GMV/UU \textbf{+4.15\%}, Ads Revenue \textbf{+3.58\%} in 7-day A/B validation, promoted to production rollout).

Despite these results, Sortify exhibits several concrete limitations that constrain its current applicability:

\begin{itemize}
  \item \textbf{Parameter oscillation not fully converged.} In the Country~A warm-start experiment retained in the main text, \texttt{ps\_ads\_wo} still exhibits a 4.5x residual pendulum effect (Section~\ref{sec:parameter-stability}). The multi-modal objective landscape created by the ads-organic trade-off is a fundamental limitation that the single-objective-with-penalties formulation can approximate but not resolve. This oscillation indicates that the system has transitioned from unbounded exploration to constrained oscillation, but has not yet converged to a single stable operating point.

  \item \textbf{Statistical uncertainty in low-traffic windows.} The strongest results (GMV +9.2\%, Orders +12.5\% in Exp-401838 R7) were observed during a low-traffic early morning window with approximately 12K impressions (Section~\ref{sec:online-ab}). While the system operated correctly, the statistical power of these estimates is lower than for rounds with typical traffic volume.

  \item \textbf{Insufficient empirical evidence for extreme structural breaks.} The retained Country~A experiments primarily demonstrate convergence in the number of LLM correction items, but do not cover a sufficiently long or severe post-warm-start lifecycle to validate handling of major structural breaks (Section~\ref{sec:llm-calibration}). This means the evidence presented for the LLM meta-controller leans more toward ``reducing ongoing calibration workload'' than ``handling extreme mutations.''

\end{itemize}

These limitations point to two directions for future work. First, replacing the single-objective-with-penalties formulation with a \textbf{multi-objective Pareto frontier} approach would allow the system to explicitly model the ads-organic trade-off as a frontier rather than a single optimum, potentially eliminating the pendulum effect. Second, developing a \textbf{confidence-weighted transfer model} that adjusts calibration aggressiveness based on observation volume would allow the system to be more conservative during low-traffic windows and more aggressive during high-traffic periods, reducing statistical risk without sacrificing convergence speed. These extensions build naturally on Sortify's existing dual-channel and Memory DB architecture, requiring modifications to the objective formulation and context assembly rather than fundamental architectural changes.

Finally, the Sortify project itself---from the foundational algorithm design and approximately 98,000 lines of engineering implementation, through 30 rigorous production experiment rounds, to the very drafting of this report---was executed entirely by a single human orchestrating a swarm of AI agents, untouched by a single line of human code or draft text. This fact stands as a silent yet deafening experiment in its own right. As explored in Section~\ref{sec:discussion}, it unveils an ``AI-generating-AI'' meta-nesting logic: an AI fleet in the creation domain birthed an industrial-grade AI hub (Sortify) in the target domain, which in turn autonomously governs highly complex business operations. This transcends mere efficiency optimization; it is a structural paradigm shift in software engineering and human exploration. When the long, arduous chasm between conceptualization and realization is fundamentally obliterated, a brutal yet fascinating new era emerges: the ceiling of creation is no longer dictated by the sheer scale of execution capacity, but solely by the cognitive penetration and orchestration artistry of the solitary decision-maker.

\newpage
\bibliographystyle{plainnat}
\bibliography{references}

\newpage
\appendix

\section{Contributors}
\label{sec:appendix-contributors}

\href{https://www.zhihu.com/people/cheng-yin-36}{\textbf{Yin Cheng}} (\texttt{yin.cheng@shopee.com}) and his AI Agents were responsible for all core work on this project:

\begin{itemize}[nosep]
  \item \textbf{Project Development}: System architecture design and full-stack engineering implementation
  \item \textbf{Algorithm Design}: Conception and iteration of the Pairwise Influence ranking algorithm
  \item \textbf{Experiment Execution}: Deployment, monitoring, and data analysis of A/B experiments
  \item \textbf{Report Writing}: Ideation, writing, and revision of this technical report
\end{itemize}

\medskip

\noindent\textbf{Acknowledgments}

\medskip

\noindent
Special thanks to Liao Zhou (\texttt{liao.zhou@shopee.com}) for his deep involvement throughout the project---providing sustained discussion, feedback, and experience transfer on experimentation methodology and development environment evolution, which played an important role in the iterative refinement of our approach.

\medskip

\noindent
We also thank the following colleagues for their support:

\smallskip
{\small
\noindent
\begin{tabular}{@{}ll@{}}
\texttt{xiyu.liang@shopee.com}   & \texttt{kailun.zheng@shopee.com} \\
\texttt{dihao.luo@shopee.com}    & \texttt{tewei.lee@shopee.com} \\
\texttt{zhangweiwei@shopee.com}  & \texttt{mark.cai@shopee.com} \\
\texttt{jian.dong@shopee.com}    & \texttt{andy.zhanggx@shopee.com} \\
\end{tabular}
}

\section{Implementation Details}
\label{sec:appendix-implementation}

\subsection{A/B Experiment Integration}

Sortify integrates with the platform's existing A/B testing infrastructure rather than deploying its own experimentation framework. Each parameter push to Redis triggers a traffic split between the control group (current production parameters) and the treatment group (Sortify's optimized parameters). The A/B report is generated by the platform's standard metrics pipeline and pulled by Sortify's one-shot pipeline in Step 1.

\textbf{Time window alignment.} All observation windows are snapped to 15-minute boundaries to align with the platform's metrics aggregation cadence. The minimum accumulation window of 3.5 hours ensures sufficient traffic volume for stable uplift estimates in most time-of-day conditions. The maximum window is capped at 3.5 hours to prevent stale data from earlier traffic patterns contaminating later observations.

\subsection{Redis Parameter Publishing}

The 7-parameter vector is serialized as JSON and written to a Redis key monitored by the sorting service. The publication is atomic---either all 7 parameters update simultaneously or none do. A rollback mechanism allows reverting to the previous parameter set by re-publishing the \texttt{PREV\_PARAMS\_JSON} stored in \texttt{yolo-state.env}.

\subsection{State File Management}

Two state files maintain continuity across pipeline executions and system restarts:

\textbf{\texttt{yolo-state.env}} --- YOLO loop state:
\begin{verbatim}
YOLO_ROUND=5
LAST_PUSH_TIME=2026-03-05T18:45:00
LAST_ONLINE_FROM=2026-03-05T15:15:00
LAST_ONLINE_TO=2026-03-05T18:45:00
PREV_PARAMS_JSON={"ps_ads_wo": 3.51, ...}
PREV_AB_REPORT_PATH=/path/to/report
\end{verbatim}

\textbf{\texttt{one-shot-latest.env}} --- Last successful one-shot handoff:
\begin{verbatim}
LATEST_RUN_TAG=20260302_143022
CONFIG_PATH=.../pdp-th-topk-order-guard-agent.next.json
BEST_SO_FAR=.../search-artifacts/*_best_so_far.json
\end{verbatim}

\subsection{Directory Structure}

Each round produces artifacts in a timestamped directory:

\begin{verbatim}
updates/
|-- memory/
|   `-- agent_memory.db              <- Master DB (shared across rounds)
|-- <run_tag>/
|   |-- online-loop.log              <- 10-step pipeline output
|   |-- search.log                   <- Optuna trial results
|   |-- ab_report.md                 <- Online A/B metrics
|   |-- llm_proposal.json            <- LLM output with evidence
|   |-- rule.publish.json            <- Published parameters
|   |-- search-logs/
|   |   `-- *_trial_table.csv        <- All 5,000 trial results
|   `-- search-artifacts/
|       `-- *_best_so_far.json       <- Best parameters found
|-- yolo-state.env
|-- yolo-event-log.md                <- Audit log of all rounds
`-- one-shot-latest.env
\end{verbatim}

\section{Notation Table}
\label{sec:appendix-notation}

\begin{table}[htbp]
\centering
\caption{TAB-APP-02. Symbol definitions.}
\label{tab:app-notation}
\small
\begin{tabular}{lll}
\toprule
\tableheaderrow
Symbol & Definition & Introduced \\
\midrule
$\pi_q(\boldsymbol{\theta})$ & Ranking permutation for request $q$ under parameters $\boldsymbol{\theta}$ & Eq.~\ref{eq:arch-15} \\
$S_{n_q}$ & Symmetric group acting on $n_q$ items & Eq.~\ref{eq:arch-15} \\
$\mathbf{S}_q(\boldsymbol{\theta})$ & Total score vector for request $q$ & Eq.~\ref{eq:arch-16} \\
$H_{qij}$ & Comparison hyperplane $x_i=x_j$ for items $i,j$ & Eq.~\ref{eq:arch-16} \\
$C_\pi$ & Ranking chamber corresponding to permutation $\pi$ & Eq.~\ref{eq:arch-16} \\
$\boldsymbol{\theta}$ & 7-dimensional sorting parameter vector & Eq.~\ref{eq:arch-04} \\
$\mathcal{F}$ & Set of all sorting factors & Eq.~\ref{eq:arch-18} \\
$S_q(i;\boldsymbol{\theta})$ & Total score of item $i$ in request $q$ & Eq.~\ref{eq:arch-17} \\
$\Delta_q(i,j;\boldsymbol{\theta})$ & Total pairwise margin for item pair $(i,j)$ & Eq.~\ref{eq:arch-17} \\
$\Delta_{q,f}(i,j;\boldsymbol{\theta})$ & Score difference contributed by factor $f$ on item pair $(i,j)$ & Eq.~\ref{eq:arch-18} \\
$Z_{qij}(\boldsymbol{\theta})$ & Total pairwise influence budget for item pair $(i,j)$ & Eq.~\ref{eq:arch-19} \\
$s_{q,f}(i,j;\boldsymbol{\theta})$ & Pairwise influence share of factor $f$ for item pair $(i,j)$ & Eq.~\ref{eq:arch-01} \\
$P_q$ & Informative / top-sensitive pair set aggregated for request $q$ & Eq.~\ref{eq:arch-03} \\
$r_{qi}$ & Rank of item $i$ in request $q$ & Eq.~\ref{eq:arch-02} \\
$w_{qij}$ & Rank-based exponential decay weight for item pair $(i,j)$ & Eq.~\ref{eq:arch-02} \\
$\tau$ & Decay rate parameter for pair weighting & Eq.~\ref{eq:arch-02} \\
$I_f(\boldsymbol{\theta})$ & Aggregate Influence Share for factor $f$ & Eq.~\ref{eq:arch-03} \\
$\mathcal{J}(\boldsymbol{\theta})$ & Composite search objective function & Eq.~\ref{eq:arch-04} \\
$\lambda_j$ & Penalty weight for constraint $j$ & Eq.~\ref{eq:arch-04} \\
$v_j(\boldsymbol{\theta})$ & Violation magnitude for constraint $j$ & Eq.~\ref{eq:arch-04} \\
$\alpha$ & Transfer model slope (Belief channel) & Eq.~\ref{eq:arch-06} \\
$\beta$ & Transfer model intercept (Belief channel) & Eq.~\ref{eq:arch-06} \\
$\eta$ & LMS learning rate (= 0.2) & Eq.~\ref{eq:arch-07} \\
$e_t$ & Transfer prediction error at round $t$ & Eq.~\ref{eq:arch-07} \\
$\Delta\beta_\text{LLM}$ & LLM intercept adjustment ($\in [-0.1, +0.1]$) & Eq.~\ref{eq:arch-09} \\
$p_j$ & Normalized violation pressure for constraint $j$ & Eq.~\ref{eq:arch-10} \\
$\delta_\text{up}$ & Penalty tightening step size (= 0.25) & Eq.~\ref{eq:arch-11} \\
$\delta_\text{down}$ & Penalty relaxation step size (= 0.08) & Eq.~\ref{eq:arch-11} \\
$m_\text{LLM}$ & LLM penalty multiplier ($\in [0.5, 2.0]$) & Eq.~\ref{eq:arch-14} \\
\bottomrule
\end{tabular}
\end{table}

\begin{table}[htbp]
\centering
\caption{Abbreviations.}
\label{tab:app-abbreviations}
\small
\begin{tabular}{ll}
\toprule
\tableheaderrow
Abbreviation & Full Form \\
\midrule
GMV  & Gross Merchandise Value \\
PDP  & Product Detail Page \\
TPE  & Tree-structured Parzen Estimator \\
LMS  & Least Mean Squares \\
YOLO & You Only Live Once (continuous pipeline mode) \\
KPI  & Key Performance Indicator \\
CTR  & Click-Through Rate \\
eCPM & Effective Cost Per Mille \\
ROI  & Return on Investment \\
\bottomrule
\end{tabular}
\end{table}

\end{document}